\pdfoutput=1
\documentclass[10pt,twocolumn,letterpaper]{article}

\usepackage[pagenumbers]{cvpr}

\makeatletter
\renewcommand\paragraph{\@startsection{paragraph}{4}{\z@}%
  {1ex \@plus .3ex \@minus .2ex}{-1em}{\normalfont\normalsize\bfseries}}
\makeatother

\usepackage{siunitx}
\usepackage{etoc}
\usepackage{microtype}
\AtEndPreamble{
  \crefname{appendix}{Sec.}{Secs.}\Crefname{appendix}{Section}{Sections}
  \crefname{subappendix}{Sec.}{Secs.}\Crefname{subappendix}{Section}{Sections}
  \crefname{subsubappendix}{Sec.}{Secs.}\Crefname{subsubappendix}{Section}{Sections}
}
\newcommand{\supp}[1]{\cref{#1}}

\definecolor{cvprblue}{rgb}{0.21,0.49,0.74}
\usepackage[pagebackref,breaklinks,colorlinks,allcolors=cvprblue]{hyperref}

\title{Dexify: Dexterous Hand--Object Retargeting Made Easy \\ with a Differentiable World Model}
\author{
  Tenghui Wang$^{1,2}$ \quad Kaiwen Song$^{3,2}$ \quad Kerui Ren$^{1,2}$ \quad Xijie Yang$^{4,2}$ \quad Xinzhe Wang$^{1,2}$ \\
  Xingyu Miao$^{2}$ \quad Yuanbo Xiangli$^{1}$ \quad Tao Lu$^{2,\dagger}$ \quad Mulin Yu$^{2,\dagger}$ \quad Feng Gao$^{1}$ \\[6pt]
  {\normalsize $^{1}$Shanghai Jiao Tong University \quad $^{2}$Shanghai Artificial Intelligence Laboratory} \\
  {\normalsize $^{3}$University of Science and Technology of China \quad $^{4}$Zhejiang University} \\[4pt]
  {\small \url{https://sanbens.github.io/dexify/}}
}

\begin{document}
\twocolumn[{%
\renewcommand\twocolumn[1][]{#1}%
\maketitle
\begin{center}
  \centering
  \vspace{-2.0em}
  \includegraphics[width=\textwidth]{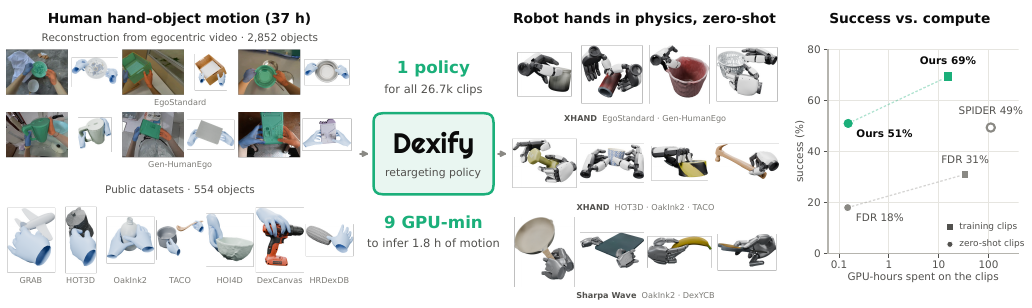}
  \captionof{figure}{\textbf{Dexify retargets human hand--object references to dexterous robot hands with one
  policy}, trained through a differentiable world model. Left: references from our egocentric-video reconstructions
  (frame and reference) and seven public datasets. Middle: zero-shot tracking of unseen objects in simulation, by our XHAND policy and by one trained the same way for the
  Sharpa Wave at its real torque limits (\cref{sec:exp_sharpa}). Right:
  success against GPU-hours on the training clips (squares, training cost) and zero-shot clips (circles, inference
  cost of the trained policy); SPIDER (hollow) optimizes each zero-shot clip instead of training.}
  \label{fig:teaser}
  \vspace{0.5em}
\end{center}
}]
{\renewcommand{\thefootnote}{\fnsymbol{footnote}}\footnotetext[2]{Corresponding authors.}}
\etocdepthtag.toc{mtmain}
\begin{abstract}
Data for dexterous robot hands is scarce. Human hand--object interactions are
a promising source: they are easy to collect at scale
from egocentric video. But existing methods for converting them into robot
data either ignore physics or are
inefficient, requiring per-clip optimization or costly model-free reinforcement learning.
We present \textbf{Dexify}, one policy that turns recorded human hand--object interactions into
physically feasible robot-hand trajectories in simulation, making them a more practical source of dexterous-hand
data.
The policy is trained by backpropagating through a differentiable world model.
Both networks see the object through its geometry relative to the hand (the world model
through the hand--object contact geometry, the policy through object inputs relative to each hand).
One mirrored module serves both hands, letting single-hand clips train the same weights that
bimanual clips use.
We train it on 37\,h of references with 3{,}406 objects, 84\% of them reconstructed from egocentric
video.
In about 15 hours on one GPU, it converts 69.4\% of its training clips and 51.0\% of zero-shot clips with unseen
objects into robot trajectories.
It outperforms a recent open-source state-of-the-art tracker with less than half of its training compute
(15 \vs 34 GPU-hours), making dexterous-hand data generation easier to scale.

\end{abstract}

\section{Introduction}
\label{sec:intro}

Human hand--object interactions are recorded at a growing scale, by motion capture and, increasingly, by
egocentric video. They are a natural source of demonstrations for dexterous robot hands,
most directly through \emph{retargeting}. Retargeting finds robot actions that manipulate the object as the
human did in a recorded clip, the \emph{reference}. Mapping hand poses to robot joints is not enough, because differences in kinematics and contact, as well as
errors in the recorded motion, can make the object slip, tip over or never leave the table. Physics-based retargeting
therefore searches in simulation for actions that reproduce the object's motion.

Most methods solve this search per demonstration, by sampling-based
optimization~\cite{pan2025spider,paliwal2026doasido} or by training one policy per
clip~\cite{li2025maniptrans,mandi2025dexmachina,zhu2026chord}, so their cost grows with the data.
\emph{Generalist trackers} instead train one policy that takes the clip to follow as input, so the training cost
is paid once rather than per clip~\cite{liu2025dextrack,lee2026fdr,luo2026dexmimic,sun2026donic}, but they
rely on model-free reinforcement learning, which is costly to scale, and have been trained on at most a few thousand
clips. Scaling retargeting beyond this regime---to tens of thousands of clips covering thousands of objects---calls for
a single policy that trains more efficiently while reliably converting both its training data and unseen clips.

A differentiable world model can offer this efficiency: SuperMimic~\cite{liao2026supermimic} trains a
humanoid tracker on a single GPU by backpropagating the return through a learned world model, a network
trained to predict the next physical state. However, when we
apply it unchanged to hand--object tracking over thousands of objects, it converts only 13.8\% of zero-shot clips,
against 51.0\% for our method (\cref{sec:exp_baselines}).
A hand's dynamics depend on where its fingers are relative to the object's surface, but SuperMimic's world model,
built for humanoids, sees no object geometry, so it predicts the object's motion poorly even on its training
clips (\supp{tab:wm_err}). On the policy side, taking an object's point cloud
as a fixed-order vector of coordinates (as in FDR~\cite{lee2026fdr}) lets the network partly memorize training objects and generalize less well to
new ones.

We present \textbf{Dexify}, one policy that retargets single-hand and bimanual interactions with a rigid object
to dexterous robot hands, trained through a differentiable world model (\cref{fig:teaser,fig:overview}).
Its central idea addresses both limitations above: \emph{each network should see an object through its geometry relative
to the hand}.
For the world model, this is the \emph{contact geometry}, the signed distances and normals from the hand to
the object's surface, which can be computed for any object and lets it predict contact dynamics accurately across
thousands of objects. For the policy, it is the object relative to each hand: an encoding of the object's point
cloud and the distances from its fingers to the surface.
Since these inputs are relative to each hand, one mirrored module can serve both hands, so single-hand clips
train the same weights that bimanual clips use.
On thousands of contact-rich clips, training through the world model also requires stronger exploration: we collect
transitions with larger, temporally correlated action noise so the world model and critic observe contact behaviors
beyond the untrained policy. With these designs, training through a differentiable world model scales to thousands of
objects, yielding a single policy that reliably converts its training clips and tracks unseen clips without further
training.

We train Dexify on 37\,h of references with 3{,}406 objects, from seven public datasets and from
\emph{EgoHOR}, the egocentric hand--object references we reconstruct from video, which contribute
2{,}852 of the objects. After about 15 hours of training on one RTX 4090, it converts 69.4\% of its training
clips into physically feasible robot trajectories in simulation and, without further training, 51.0\% of
zero-shot clips from held-out objects, categories and datasets. On the same data in our simulator,
FlashDexRetarget (FDR)~\cite{lee2026fdr}, a recent open-source generalist tracker trained with model-free
reinforcement learning, converts 31.1\% and 18.0\%, although it trains 2.2 times as long.
SPIDER~\cite{pan2025spider}, which optimizes every zero-shot clip separately by sampling, converts 49.4\% of
them, about as many as our policy, at a cost of 111 GPU-hours, about seven times our whole training.
Zero-shot success also grows with the training data, and EgoHOR helps most where public data is scarce
(\cref{sec:exp_scaling}); every group of components matters (\cref{sec:exp_ablation}).
Trained unchanged on a second robot hand (Sharpa Wave), Dexify reaches 70.3\% zero-shot success at its real torque
limits (\cref{sec:exp_sharpa}), and trajectories converted by Dexify yield higher downstream imitation-learning success
than those from FDR and reference replay (\cref{sec:exp_bc}).

In summary, our contributions are:
\begin{itemize}
  \item Dexify, one retargeting policy for single-hand and bimanual interactions, trained through a
  differentiable world model. Both networks see the object's geometry relative to the hand,
  scaling world-model-based retargeting to thousands of objects and unseen shapes
  (\cref{sec:method}).
  \item A data pipeline that reconstructs egocentric hand--object clips (EgoHOR) and converts them, together with seven
  public datasets, into 37\,h of robot references over 3{,}406 objects (\cref{sec:data}); adding EgoHOR improves
  zero-shot tracking on held-out categories and when public data is scarce (\cref{sec:exp_scaling}).
  \item Experiments showing that Dexify converts 69.4\% of training and 51.0\% of zero-shot clips into physically
  feasible trajectories in simulation, far exceeding model-free RL at under half the compute and matching per-clip
  optimization at a seventh of its compute; it also transfers unchanged to Sharpa Wave (70.3\% zero-shot at real torque
  limits) and improves downstream imitation learning (\cref{sec:exp}).
\end{itemize}
We will release the code, data pipeline and checkpoints.

\section{Related Work}
\label{sec:related}

\paragraph{Physics-based dexterous retargeting and tracking.}
Physics-based retargeting searches in simulation for actions that reproduce the object's motion, mostly per
demonstration: by sampling-based optimization in SPIDER~\cite{pan2025spider} and Do as I Do~\cite{paliwal2026doasido},
a curriculum of quasi-physical simulators in QuasiSim~\cite{liu2024quasisim}, or per-clip RL in
ManipTrans~\cite{li2025maniptrans}, DexMachina~\cite{mandi2025dexmachina}, CHORD~\cite{zhu2026chord},
ConTrack~\cite{liang2026contrack} and AdaDex~\cite{kuang2026adadex}.
SPIDER is an order of magnitude faster than per-clip RL, yet still searches every clip from scratch; the cost of all these methods grows with the number of
demonstrations.
Generalist trackers instead train one reference-conditioned policy over many demonstrations
(DexTrack~\cite{liu2025dextrack}, Dexplore~\cite{xu2025dexplore}, BiDexHD~\cite{zhou2026bidexhd},
FDR~\cite{lee2026fdr}, DexMimic~\cite{luo2026dexmimic}, DONIC~\cite{sun2026donic}), several of which also track
unseen sequences or objects without fine-tuning.
FDR, the closest open-source generalist tracker to our setting, trains one off-policy agent on about 1.5k bimanual XHAND clips and, in its own
study, outperforms per-clip RL and sampling-based optimization.
These trackers are model-free, train on up to a few thousand clips as far as reported, and mostly handle only
single-hand or only bimanual clips; we train one policy for both on thousands of objects.
Prior methods describe the object to the policy by a pretrained point-cloud embedding (DexTrack), basis point
sets (ManipTrans, AdaDex, DONIC) or a point cloud with signed hand--object distances (FDR); our policy instead trains a
point-order-invariant PointNet encoder end-to-end with the policy, recomputed at every step from the object's points in each
hand's frame (\cref{sec:policy}).

\paragraph{Learned world models and analytic policy gradients.}
Character and humanoid controllers track large motion libraries with model-free
RL~\cite{peng2018deepmimic,luo2023perpetual,liao2025beyondmimic,luo2025sonic}, at a cost that grows with the
library (SONIC: 128 GPUs for a week). Model-based RL lowers this cost by learning the dynamics, but
MBPO~\cite{janner2019trust}, DreamerV3~\cite{hafner2025dreamerv3} and TD-MPC2~\cite{hansen2024tdmpc2} learn from or
plan in a learned model and estimate policy gradients from sampled returns or a value function, not by
differentiating through the dynamics.
Analytic policy gradients backpropagate the return through a differentiable simulator
(SHAC~\cite{xu2022accelerated}, $\nabla$Mimic~\cite{lu2026nablamimic}) or a learned world model
(SuperTrack~\cite{fussell2021supertrack}, ControlVAE~\cite{yao2022controlvae}, SuperMimic~\cite{liao2026supermimic}).
These methods control bodies that mostly contact the ground, and their learned world models see only the body's
state; SuperMimic leaves object interaction to future work.
Learned simulators that take object geometry as input, such as particle- and mesh-based graph
networks~\cite{li2019dpinet,sanchezgonzalez2020gns,allen2023fignet}, generalize to new shapes but model every
particle or mesh face, which makes them costly to train a policy through.
We bring policy learning by backpropagation through time (BPTT) in a learned world model to contact-rich
hand--object tracking over thousands of objects; our world model captures hand--object contact dynamics across diverse
shapes through the contact geometry and predicts a compact hand--object state, cheap enough to train a policy through (\cref{sec:wm}).

\paragraph{Learning from human video.}
Human videos are increasingly used for robot learning: prior work reconstructs hand--object trajectories from
them~\cite{chen2026videomanip,wang2026egoinfinity}, turns them into robot demonstrations or edited robot
videos~\cite{liu2026egoengine,guo2026roboedit}, or records them at scale with tracked hands~\cite{hoque2026egodex}.
Hand--object reconstruction from video has advanced through category-agnostic and prior-guided
methods~\cite{fan2024hold,ye2023diffhoi,liu2025easyhoi}.
We combine such components into a reconstruction pipeline and use its output as references to train one
retargeting policy for all clips, rather than one per manipulation task.

\section{Constructing Hand--Object References}
\label{sec:data}

Dexify takes human hand--object reference clips as input and converts them into robot trajectories in simulation
(\cref{sec:method}). To train it across thousands of objects, we construct 37\,h of such clips (\cref{tab:data}). Each
clip records human hands manipulating a rigid object, with MANO parameters~\cite{romero2017mano} for the hands and a
mesh with a 6D trajectory for the object. Of the 37\,h, 21\,h come from seven public hand--object datasets (the
\emph{public data} below), which require mesh and 6D pose annotations for each object and cover 554 objects in total.
To scale to thousands of objects, we add 16\,h of EgoHOR, references of 2{,}852 objects that we reconstruct from egocentric video
(\cref{sec:egorecon}).
\label{sec:data_public}
The public datasets contribute single-hand and bimanual clips, whereas all EgoHOR clips track two hands.
All clips from both sources are then mapped to the robot hand by \emph{kinematic retargeting}
and filtered in the same way (\cref{sec:processing}). This mapping ignores physics; Dexify then tracks the clips under
physics.

\begin{table}[t]
  \centering
  \footnotesize
  \setlength{\tabcolsep}{3pt}
  \caption{\textbf{Training references} (zero-shot sets excluded). EgoHOR: our reconstructions from the
  EgoStandard and Gen-HumanEgo videos (\cref{sec:egorecon}).}
  \label{tab:data}
  \begin{tabular}{@{}lcrrr@{}}
    \toprule
    Source & Hands & Clips & Hours & Objects \\
    \midrule
    HOT3D~\cite{banerjee2025hot3d}     & 1--2 & 1{,}464 & 2.73 & 27  \\
    GRAB~\cite{taheri2020grab}         & 2    & 377     & 1.01 & 44  \\
    DexCanvas~\cite{xu2025dexcanvas}   & 1    & 7{,}864 & 7.93 & 21  \\
    HRDexDB~\cite{lim2026hrdexdb}      & 1    & 1{,}119 & 3.97 & 56  \\
    OakInk2~\cite{zhan2024oakink2}     & 1--2 & 996     & 1.52 & 61  \\
    TACO~\cite{liu2024taco}            & 2    & 3{,}025 & 2.21 & 133 \\
    HOI4D~\cite{liu2022hoi4d}          & 1    & 468     & 1.59 & 212 \\
    \textit{Public datasets}           &      & 15{,}313 & 20.95 & 554 \\
    \midrule
    EgoStandard~\cite{lightwheel2026egosuite}   & 2 & 8{,}451 & 12.33 & 1{,}862 \\
    Gen-HumanEgo~\cite{genrobot2026genhumanego} & 2 & 2{,}984 & 4.00  & 990 \\
    \textit{EgoHOR}                   &      & 11{,}435 & 16.33 & 2{,}852 \\
    \midrule
    Total                              &      & 26{,}748 & 37.28 & 3{,}406 \\
    \bottomrule
  \end{tabular}
\end{table}

\subsection{Reconstruction from egocentric video}
\label{sec:egorecon}

\begin{figure*}[t]
  \centering
  \includegraphics[width=\textwidth]{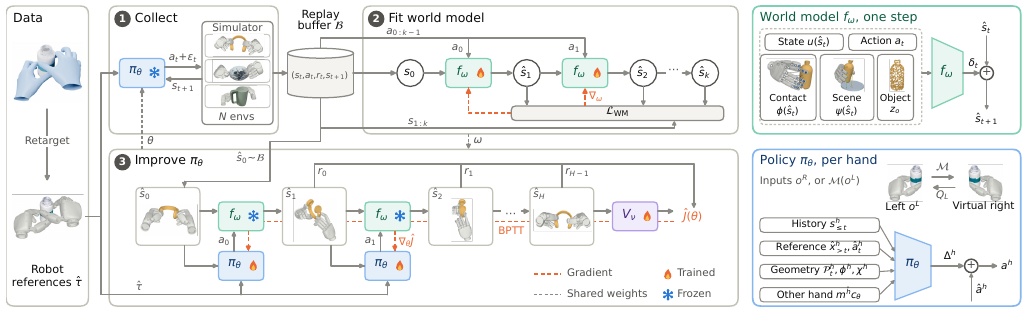}
  \caption{\textbf{Overview.}
  The policy~$\pi_\theta$ is trained through a learned world model~$f_\omega$ (the loop of
  SuperMimic~\cite{liao2026supermimic}, drawn unrolled): (1)~collect transitions into a replay buffer~$\mathcal{B}$;
  (2)~fit $f_\omega$ on $k$-step windows of~$\mathcal{B}$; (3)~improve~$\pi_\theta$ by BPTT over $H$ steps imagined
  with the frozen~$f_\omega$, with a critic~$V_\nu$ beyond~$H$.
  Right: our world model, which sees contact and scene geometry~$\phi,\psi$ (\cref{fig:wm_geom}) and an object
  code~$z_o$, and our policy, one module shared by both hands, the left mirrored.}
  \label{fig:overview}
\end{figure*}

EgoHOR extends the training data beyond the objects of the public datasets: egocentric video records whatever
objects people handle but provides no object meshes or poses, so we reconstruct them.
Our sources are EgoStandard~\cite{lightwheel2026egosuite} and
Gen-HumanEgo~\cite{genrobot2026genhumanego}, egocentric recordings with stereo video and tracked hand
keypoints (\supp{sec:supp_recon}).
In segments where the hands interact with one rigid object (selected by a vision--language model or by the
recordings' subtask annotations), we segment the object (SAM3~\cite{carion2025sam3}, from a point prompt of
the vision--language model), estimate metric depth (Fast-FoundationStereo~\cite{wen2026fastfoundationstereo}),
track points on the object (TAPNext++~\cite{jung2026tapnextpp}), and generate a metric mesh
(Fast-SAM3D~\cite{feng2026fastsam3d}); the vision--language model also estimates the object's mass. We then fit MANO to the hand keypoints, and the object's 6D trajectory
to the tracked points, silhouette and depth. Automatic checks reject failed reconstructions.
The reconstructed references, sampled at about \qty{10}{\hertz}, have no ground truth and contain artifacts such
as holes in the meshes and hands that penetrate the object. Where depth drift or tracking errors move the object
with no hand near it, we move the object toward the hand before the kinematic retargeting
(\supp{sec:supp_processing}).
\Cref{fig:supp_recon} shows four EgoHOR clips with our policy's zero-shot tracking.

\subsection{Kinematic retargeting and filtering}
\label{sec:processing}

Tracking the reference clips from both sources with robot hands under physics (\cref{sec:method}) requires three
things they lack: robot joint targets, a scene around the object, and a physically reproducible motion. Kinematic retargeting gives the joint targets, support surfaces give the scene, and filtering removes physically implausible or unstable clips. The human hands are mapped to the robot hand with FDR's
kinematic retargeting, providing the reference joint targets (\cref{sec:task}).
No source provides the scene around the object, so as a minimal scene we place a support surface wherever
the object rests in the reference (\supp{sec:supp_scene}).
Filtering removes references unsuitable for physics-based tracking: we discard clips where the object starts unsupported, moves untouched, or fails retargeting.

\section{Method}
\label{sec:method}

Dexify trains one policy to track the reference clips of \cref{sec:data} using analytic gradients through a learned
differentiable world model (\cref{fig:overview}). While the outer training loop follows
SuperMimic~\cite{liao2026supermimic} (introduced with the task in \cref{sec:task}), moving from a single humanoid on
flat ground to two dexterous hands manipulating thousands of objects introduces fundamentally different challenges:
contact dynamics depend on each object's shape, one policy must handle diverse and unseen shapes across both hands, and
analytic gradients become fragile under dense, discontinuous contact. We address these challenges across every part of
the loop. \Cref{sec:wm} builds a geometry-aware world model that captures hand--object and scene contact for arbitrary
shapes; \cref{sec:policy} designs a hand-relative policy with a mirrored module shared by both hands; and
\cref{sec:training} makes policy optimization reliable on contact-rich clips.

\subsection{Preliminaries}
\label{sec:task}
\label{sec:prelim}

\paragraph{Task.}
The policy controls two robot hands in simulation so that the object follows a reference clip.
A kinematically retargeted reference clip of \cref{sec:data} is $\tau_i=\{\hat{x}_{i,t}\}_{t=0}^{T_i}$: per frame, each
hand's MANO wrist, joint and fingertip positions and robot joint targets $\hat{q}_t$, and the object pose
$(\hat{p}_t,\hat{R}_t)$.
Two XHANDs~\cite{robotera_xhand} track the clips in MuJoCo-Warp~\cite{mujocowarp}, each with a floating 6-DoF wrist,
which stands in for a robot arm (\supp{sec:supp_sim}), and 12 finger joints, driven by PD targets; the scene consists
of the support surfaces of \cref{sec:processing}. In a single-hand clip the other hand is idle (\supp{sec:supp_sim}), so single-hand and bimanual clips run in the
same bimanual simulation under one policy $\pi_\theta$. The policy outputs a residual on the reference PD target,
\begin{equation}
  a^h_t=\tanh\!\big(\operatorname{atanh}(\hat{a}^h_t)+\Delta^h_t\big),
  \label{eq:residual}
\end{equation}
where $\hat{a}^h_t$ is hand $h$'s normalized reference target and $\Delta^h_t$ the policy's output
(\cref{eq:mirror}), so the untrained policy replays the reference. During training, episodes terminate when the object or
a fingertip diverges from the reference.

\paragraph{Training through a world model.}
Dexify trains $\pi_\theta$ with the loop of SuperMimic~\cite{liao2026supermimic}, which gives the policy analytic gradients from a
learned, differentiable world model instead of the simulator. It alternates three phases (\cref{fig:overview}). (1)~The policy collects
simulator transitions into a replay buffer $\mathcal{B}$. (2)~A world model $\hat{s}_{t+1}=f_\omega(\hat{s}_t,a_t)$
(\cref{sec:wm}) is fit on $k$-step windows of $\mathcal{B}$ using its own rollouts,
\begin{equation}
  \mathcal{L}_{\text{WM}}(\omega)=\mathbb{E}_{\mathcal{B}}\textstyle\sum_{j=1}^{k}\lVert\hat{s}_j-s_j\rVert^2,
  \quad \hat{s}_0=s_0 .
  \label{eq:wm}
\end{equation}
(3)~$\pi_\theta$ is improved by BPTT over $H$ imagined steps from buffer states, with the reward $r$ of \cref{eq:reward} and a critic $V_\nu$ for the return beyond $H$,
\begingroup\small
\begin{equation}
  \hat{J}(\theta)=\mathbb{E}\big[\textstyle\sum_{t=0}^{H-1}\gamma^t\hat{c}_t\,r(\hat{s}_{t+1},a_t)
    +\gamma^H\hat{c}_H V_\nu(\hat{s}_H)\big],
  \label{eq:bptt}
\end{equation}
\endgroup
where $\hat{c}_t$ is a differentiable estimate of the probability that the episode has not terminated before
step $t$; the critic is fit to TD($\lambda$) returns of the imagined rollouts, as in generalized advantage estimation
(GAE)~\cite{schulman2016gae}. Rewards and terminations are computed from predicted states via differentiable kinematics, so the policy receives their gradients.

\subsection{Geometry-aware world model}
\label{sec:wm}

The world model predicts how the hands and the object move from one step to the next, and the policy is trained on
these predictions (\cref{sec:task}). When the fingers touch the object, the resulting motion depends on where they meet
the object's surface---information absent from the joint angles and object pose alone. Alongside the state and actions,
we therefore feed the world model the contact geometry (each hand part relative to the object's surface), scene geometry and an object code, all computed from the mesh and scene without per-object parameters.

\paragraph{Inputs and outputs.}
Our $f_\omega$ predicts the next state from the state $\hat{s}_t=(q_t,\dot{q}_t,p_t,R_t,v_t,\omega_t)$ (both hands' joint positions and velocities, the object's pose and velocities) by applying an increment predicted by a network
$\delta_\omega$,
\begin{equation}
  f_\omega(\hat{s}_t,a_t)=\hat{s}_t\oplus\delta_\omega\big(u_t,\,a_t,\,\phi_t,\,\psi_t,\,z_o\big),
  \label{eq:wmstep}
\end{equation}
where $u_t=u(\hat{s}_t)$ maps the state to hand-relative features, $a_t$ holds both hands' actions
(\cref{eq:residual}), $\phi_t,\psi_t$ are the contact and scene geometry of $\hat{s}_t$
(\cref{eq:phi}), $z_o$ is the object code (below), and $\oplus$ applies the increments (\supp{sec:supp_wm}).

\begin{figure}[t]
  \centering
  \includegraphics[width=\linewidth]{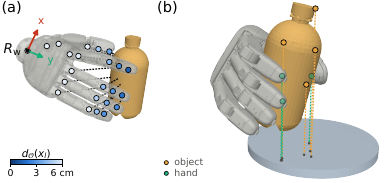}
  \caption{\textbf{Contact and scene geometry}.
  (a)~Contact geometry $\phi$ as the hand approaches: points $x_l$ inside the translucent hand, colored by
  $d_{\mathcal{O}}$ and, within the clipping range, joined to their closest surface point; normals are expressed
  in the wrist frame $R_{\text{w}}$.
  (b)~Scene geometry $\psi$ after lift-off: distances $d_{\mathcal{S}}$ from points $y_k$ on the object, palm and
  fingertips (a subset) to the support surface.}
  \label{fig:wm_geom}
\end{figure}

\paragraph{Contact and scene geometry.}
The world model receives the hand--object contact geometry $\phi$ and the scene geometry $\psi$
(\cref{fig:wm_geom}),
\begin{equation}
  \begin{aligned}
    \phi(\hat{s}) &= \big[\operatorname{clip}(d_{\mathcal{O}}(x_l)),\;R_{\text{w}}^{\top}\nabla d_{\mathcal{O}}(x_l)\big]_{l},\\
    \psi(\hat{s}) &= \big[\operatorname{clip}(d_{\mathcal{S}}(y_k))\big]_{k},
  \end{aligned}
  \label{eq:phi}
\end{equation}
where $x_l$ are points on each hand's palm, fingertips and finger links, $y_k$ points on the object and
on the palms and fingertips, $d_{\mathcal{O}}$ and $d_{\mathcal{S}}$ the signed distance fields of the object
and of the static scene $\mathcal{S}$ (the support surfaces), clipped to a band
around the surface, and $R_{\text{w}}$ the wrist orientation of the hand of $x_l$.
$\phi$ tells the world model where the fingers are relative to the object's surface, also before contact, and $\psi$
tells it whether the object rests or is lifted. Both are recomputed from the predicted state at every
imagined step and differentiable, so the policy gradient flows through them.

\paragraph{Object code.}
To give the world model the object's global shape and physical properties (such as mass), we compute an object code
$z_o$ from its surface points (through a PointNet~\cite{qi2017pointnet}) and physical parameters (\supp{sec:supp_wm}).
Unlike a learned per-object embedding, $z_o$ needs no per-object parameters and is defined for any object.

\subsection{Hand-relative policy}
\label{sec:policy}

Trained through the world model (\cref{sec:wm}), the policy $\pi_\theta$ converts both training and unseen reference
clips into physically feasible robot trajectories in simulation. At every step, it reads each hand's state, the upcoming reference
frames and the object, and outputs a residual $\Delta^h_t$ on that hand's reference action (\cref{eq:residual}). To
generalize to unseen shapes, it sees the object relative to each hand through inputs that need no per-object
parameters. Because these inputs are hand-relative, one mirrored module serves both hands.
Hand $h$ observes
\begin{equation}
  o^h_t=\big(s^h_{\le t},\;\hat{x}^h_{>t},\;\hat{a}^h_t,\;\mathcal{P}^h_t,\;\phi^h,\;\chi^h\big),
  \label{eq:obs}
\end{equation}
where, with poses and points in the hand's wrist frame, $s^h_{\le t}$ is a short history of joint states,
previous actions and the object's pose and velocities; $\hat{x}^h_{>t}$ the upcoming reference states (next 10 frames);
$\hat{a}^h_t$ the current-step reference action (\cref{eq:residual}); $\mathcal{P}^h_t$ the hand-relative point-cloud
encoding; $\phi^h$ the contact geometry; and $\chi^h$ local probes, the object's
signed distances on small grids around the palm and fingertips. From these, one mirrored module shared by both
hands outputs the residual $\Delta^h_t$ (\cref{eq:mirror}).

\paragraph{Hand-relative point-cloud encoding.}
FDR feeds the object's point cloud to the policy as a flat vector in a fixed point order, which lets the network
partly memorize each object; we instead encode it with a PointNet, $\mathcal{P}^h_t=\max_m h'([p_m,\Delta p_m])$, where $p_m$ are the object's surface points in the wrist frame (a farthest-point subset), $\Delta p_m$ each
point's displacement to its position at the next reference pose, and $h'$ a per-point MLP trained
end-to-end through BPTT. The encoding is invariant to point order and, recomputed at every
step, carries the object's pose relative to the hand and its displacement to the target pose.

\paragraph{Mirrored shared hand module.}
Because the robot's left hand is the exact mirror image of its right, we reflect the left hand's inputs into those of
a virtual right hand and run one shared module for both. This lets both hands share training data, which matters
because most single-hand clips are right-handed: a policy that takes both hands' inputs side by side, or one policy
per hand~\cite{lee2026fdr}, learns the left hand only from clips that use it, so the many right-handed clips do not
help.
The residual of hand $h$, with $\bar{h}$ the other hand, is
\begingroup\small
\begin{equation}
  \Delta^h_t = Q_h\odot\pi_\theta\big(\mathcal{M}_h(o^h_t),\, e_h,\,
    m^{\bar{h}}_t\,c_\theta(\mathcal{M}_{\bar{h}}(o^{\bar{h}}_t))\big),
  \label{eq:mirror}
\end{equation}
\endgroup
where $\mathcal{M}_h$ is the identity for the right hand and, for the left, a reflection of the world
through a vertical plane applied to all of its inputs $o^h_t$; $\odot$ denotes element-wise multiplication and
$Q_h\in\{\pm1\}^{18}$ flips the signs of the mirrored joints back; $e_h$ is a one-hot hand identity,
$c_\theta$ a learned summary of the other hand, and $m^{\bar{h}}_t\in\{0,1\}$ whether it is active.
A single-hand clip thus trains the same module used in bimanual clips, and an idle hand does not affect the active one.

\subsection{Policy optimization}
\label{sec:training}

The policy is improved by backpropagating the total expected reward (return) through an $H$-step imagined rollout
(\cref{eq:bptt}). Because the rollout only looks $H$ steps ahead, this return splits into two parts: the reward scores
each of the $H$ imagined steps, and the critic predicts the remaining cumulative reward beyond step $H$. Both the
critic and the world model that generates the rollout are trained on transitions collected by the policy with
exploration noise. Below we detail our reward, critic, and exploration designs for contact-rich manipulation across
thousands of objects. All hyperparameters are in
\supp{tab:hparams_task,tab:hparams_learn}.

\paragraph{Reward.}
The policy maximizes the return of a tracking reward adapted from FDR's. Per hand, the base reward $r^{\text{FDR}}$
tracks the object's surface points ($r_{\text{pcd}}$), fingertip--surface distances ($r_{\text{int}}$), and hand poses
with retargeted joints ($r_{\text{imit}}$), alongside action regularization ($r_{\text{reg}}$). We adjust this
base reward and add two groups of terms. First, to provide smooth analytic gradients (\cref{eq:bptt}), we replace
FDR's kernel $\exp(-e/\sigma)$, whose gradient does not vanish at zero error and flips sign across it, by a
squared kernel $\exp(-e^2/\sigma^2)$ for the point-cloud and fingertip-position terms, add an action-rate penalty in
$r_{\text{reg}}$, and omit the termination penalty. Second, for slender or axially symmetric objects, rotations about
the long axis move surface points very little and are weakly penalized by $r_{\text{pcd}}$, so we add
$r_{\text{rot}}=\exp(-\angle(R_t,\hat{R}_t))$ on the object's geodesic orientation error. Third, since tracked objects still jitter, we track reference object and wrist velocities ($r_{\text{vel}}$) and penalize
wrist-action changes ($r_{\Delta\text{w}}$) and action second differences ($r_{\text{jerk}}$).
Per hand,
\begin{equation}
  r_t = r^{\text{FDR}}_t + w_{\text{rot}}r_{\text{rot}} + w_{\text{vel}}r_{\text{vel}}
        + w_{\Delta\text{w}}r_{\Delta\text{w}} + w_{\text{jerk}}r_{\text{jerk}},
  \label{eq:reward}
\end{equation}
and the step reward is the mean over the active hands. All terms and
weights are in \supp{sec:supp_reward,tab:hparams_task}.

\paragraph{Critic.}
The critic $V_\nu$ estimates the return beyond step $H$ for the full bimanual state from both hands' features
$\zeta^h_t$ (without mirroring): each hand's world-model state features, its tracking errors, the upcoming reference
frames and contact geometry at the palm and fingertips (\supp{sec:supp_critic}).
In single-hand clips, the critic receives the activity flags $m^h_t$ of \cref{eq:mirror} and zeros for the idle
hand. Besides imagined TD($\lambda$) returns, we also fit the critic to TD($\lambda$) returns on short windows of
recent simulator transitions (\supp{tab:hparams_learn}).

\paragraph{Exploration.}
Because the world model and critic learn from the replay buffer, exploration must expose them to diverse hand--object
contacts. Instead of SuperMimic's small, uncorrelated Gaussian noise ($\sigma=0.1$), we collect data with larger,
temporally correlated action noise---implemented across four noise levels $\sigma\in\{0,0.2,0.4,0.6\}$ per block of
environments~\cite{singla2024sapg}, with the critic's real-return windows restricted to $\sigma\le0.2$ (the
return filter)---and low-pass filter the finger targets (\supp{tab:hparams_learn}).

\section{Experiments}
\label{sec:exp}

We first evaluate tracking accuracy, compute efficiency, data scaling, and design choices against learned trackers and
per-clip optimization (\cref{sec:exp_baselines,sec:exp_scaling,sec:exp_ablation}). We then test cross-embodiment
transfer to a second robot hand (\cref{sec:exp_sharpa}), downstream closed-loop imitation learning (\cref{sec:exp_bc}),
real-robot replay (\cref{sec:exp_real}), and qualitative results with failure modes (\cref{sec:exp_qual}).

\subsection{Setup}
\label{sec:setup}

\paragraph{Training.}
Unless noted, all learned methods train on the same 37\,h of references (26{,}748 clips) in
the same simulator, with four times the XHAND's finger torque limits (\emph{4$\times$}).
We raise the limit because the references assume human strength. At its real limits (\emph{1$\times$}), the
XHAND's grip force is at most about 80\,N~\cite{robotera_xhand}, about a quarter of the average adult woman's
grip strength (about 300\,N~\cite{dodds2014grip}); at 4$\times$ it is about the same, and still below the average adult
man's (about 500\,N).
The supplement repeats the data experiments (\cref{sec:exp_scaling}), the baseline comparison and the ablations at
1$\times$, and gives full per-set results and details for the second hand (\cref{sec:exp_sharpa}) and downstream
imitation learning (\cref{sec:exp_bc}).
Our policy and its variants train for 40k iterations on one RTX 4090, with two or three seeds.

\paragraph{Evaluation sets.}
\label{sec:splits}
We evaluate in-domain on 863 \emph{training clips} (563 public, 300 from EgoHOR)
and out-of-distribution on seven \emph{zero-shot} sets, named after their source with the suffix -ZS. They hold
out object categories of OakInk2 (bimanual and single-hand clips as separate sets, OakInk2-ZS and OakInk2-ZS-1H),
TACO and the two EgoHOR sources (EgoStd-ZS, GenHE-ZS), objects of HOT3D, or all of DexYCB~\cite{chao2021dexycb}; \supp{sec:supp_zs} lists them and our leakage checks.
The training clips are the data to retarget: every one the policy tracks yields a robot trajectory that is
physically feasible in simulation, so success on them measures how much of the data is converted, the main goal of
retargeting. Zero-shot success measures whether new data can be converted without further training. A clip is \emph{converted} when its rollout passes the criterion below.

\paragraph{Metrics and protocol.}
\label{sec:metrics}
Each clip is rolled out from its first frame without early termination. The main
metric, SR$_{\text{MT}}$, is the strictest criterion of ManipTrans~\cite{li2025maniptrans} and FDR: mean object
position and rotation errors within \qty{3}{cm} and \ang{30}, and mean fingertip and hand-keypoint errors of
every active hand within \qty{6}{cm} and \qty{8}{cm} (\supp{sec:supp_eval}). We also report the normalized
return (the episode return divided by the largest attainable one),
which needs no threshold, and the training GPU-hours. Tables report the mean over seeds of three evaluations of
each run's final checkpoint (SPIDER: one run per clip); seed ranges are in the supplement.

\subsection{Comparison with baselines}
\label{sec:exp_baselines}

All baselines run in our simulator with the same hands, physics and references, and all methods are evaluated on
fresh rollouts of their final policy (\cref{tab:baselines}).
\emph{FDR}~\cite{lee2026fdr} is the strongest open-source model-free generalist tracker we found, and
outperforms per-clip methods in its own study.
We reimplement it from its released code and validate the reimplementation in FDR's own bimanual setting,
where it matches the original implementation (\supp{sec:supp_baselines}).
We train it for 600M steps
with three seeds.
\emph{SuperMimic}~\cite{liao2026supermimic}, the world-model
loop we build on, runs with its original recipe for 40k iterations, about as many GPU-hours as ours
(\supp{sec:supp_supermimic}). \emph{SPIDER}~\cite{pan2025spider}
optimizes every test clip by sampling-based model-predictive control on our simulator's model; we run
it on all zero-shot clips. \emph{Reference replay} plays back the retargeted joint targets without learned correction.

\begin{table}[t]
  \centering
  \small
  \setlength{\tabcolsep}{3pt}
  \caption{\textbf{Comparison with baselines}, all in our simulator: SR$_{\text{MT}}$ (\%),
  normalized return and training GPU-hours (RTX 4090); zero-shot: mean of the seven sets (per set:
  \supp{tab:base_sets}). $^{\dagger}$Test-time optimization of the zero-shot clips.}
  \label{tab:baselines}
  \begin{tabular}{@{}l cc cc r@{}}
    \toprule
    & \multicolumn{2}{c}{Training clips} & \multicolumn{2}{c}{Zero-shot} & \\
    \cmidrule(lr){2-3} \cmidrule(lr){4-5}
    Method & SR$_{\text{MT}}$ & Return & SR$_{\text{MT}}$ & Return & GPU-h \\
    \midrule
    Ref.\ replay                         & 19.7 & 0.587 & 11.5 & 0.542 & -- \\
    SPIDER~\cite{pan2025spider}          & --   & --    & 49.4 & --    & 111$^{\dagger}$ \\
    SuperMimic~\cite{liao2026supermimic} & 28.2 & 0.501 & 13.8 & 0.426 & 13.9 \\
    FDR~\cite{lee2026fdr}                & 31.1 & 0.492 & 18.0 & 0.425 & 33.9 \\
    Ours & \textbf{69.4} & \textbf{0.688} & \textbf{51.0} & \textbf{0.614} & 15.3 \\
    \bottomrule
  \end{tabular}
\end{table}

\paragraph{Better than learned trackers.}
Our policy outperforms FDR on the training clips and on every zero-shot set (\cref{tab:baselines}): 69.4 \vs
31.1\% on training clips and 51.0 \vs 18.0\% zero-shot, although FDR trains 2.2 times as long. FDR's zero-shot success barely moves with more steps, from 15.7\% at 100M steps to 18.0\% at 600M
(\cref{fig:scaling}b); at about our training time (300M steps, 17.0 GPU-hours) it reaches 17.1\%.
SuperMimic reaches only 13.8\% zero-shot at about our compute, so the world-model loop alone does not explain our
results.

\paragraph{Against per-clip optimization.}
SPIDER, although it optimizes every test clip, converts about as many zero-shot clips as our policy
(49.4 \vs 51.0\%; \cref{tab:baselines}): SPIDER is ahead on the public sets (55.5 \vs 52.5\% on average), our policy on the
EgoHOR sets (47.2 \vs 34.2\%; per set in \supp{tab:spider_sets}). SPIDER's search costs 111 GPU-hours for these clips, about seven times our whole training.
None of this search is reused for the next clip, whereas our policy converts a new clip at the cost of one
rollout. At the real torque limits (1$\times$), SPIDER converts more zero-shot clips than our policy
(42.4 \vs 31.0\%; \supp{tab:sets_t1}), at about seven times our training compute. Its search finds grasps that
the weak fingers can hold.

\subsection{Effect of the amount and source of training data}
\label{sec:exp_scaling}
\label{sec:exp_ego}

\begin{figure}[t]
  \centering
  \includegraphics[width=\linewidth]{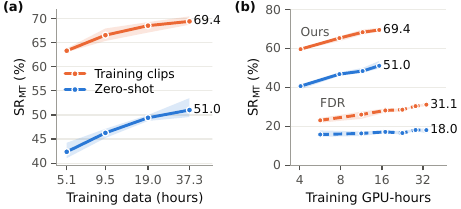}
  \caption{\textbf{Data scaling}. (a)~SR$_{\text{MT}}$ on the zero-shot sets and on the 863 training clips of the full
  data (smaller subsets train on only some of them), against the amount of training data. (b)~SR$_{\text{MT}}$
  against training GPU-hours with all data: ours at 10k--40k iterations, FDR every 100M steps. Bands: seed range.}
  \label{fig:scaling}
\end{figure}

We vary the amount and the source of the training data. Subsets are nested, and larger ones add new objects
(\supp{sec:supp_scaling}). First, we train on 12.5, 25, 50 and 100\% of all data (\cref{fig:scaling}). Second, to test if EgoHOR helps, we add none, 12.5\% or all of it to all or part of the public
data (\cref{tab:ego}).

\begin{table}[t]
  \centering
  \small
  \setlength{\tabcolsep}{3.5pt}
  \caption{\textbf{Effect of adding EgoHOR}. SR$_{\text{MT}}$ (\%) when training on a share of the
  public data and of EgoHOR. Zero-shot: means over the five public, two EgoHOR and all seven sets. EgoStd-ZS, GenHE-ZS,
  training clips, and 1$\times$ results in \supp{tab:supp_ego}.}
  \label{tab:ego}
  \begin{tabular}{@{}rr ccc@{}}
    \toprule
    Public & EgoHOR & Public ZS & EgoHOR ZS & ZS mean \\
    \midrule
    100\% & 100\%  & \textbf{52.5} & \textbf{47.2} & \textbf{51.0} \\
    100\% & 12.5\% & 51.4 & 42.6 & 48.9 \\
    100\% & 0\%    & 51.6 & 39.7 & 48.2 \\
    \midrule
    50\%  & 100\%  & 50.5 & 44.2 & 48.7 \\
    50\%  & 0\%    & 47.2 & 37.5 & 44.5 \\
    \midrule
    25\%  & 100\%  & 48.5 & 43.5 & 47.1 \\
    25\%  & 0\%    & 43.1 & 32.8 & 40.2 \\
    \bottomrule
  \end{tabular}
\end{table}

\paragraph{Conversion rate grows with data.}
Zero-shot success rises with every doubling of the data, from 42.4\% to 46.3, 49.4 and 51.0\%
(\cref{fig:scaling}a), and on the 863 training clips of the full data from 63.3\% to 69.4\%.

\paragraph{Training on EgoHOR helps on its held-out categories.}
With all public data (\cref{tab:ego}), EgoHOR raises zero-shot success on the EgoHOR sets from 39.7\% to 47.2\%
($+7.8$ points on EgoStd-ZS, $+7.2$ on GenHE-ZS), and the public sets change little (51.6 and 52.5\%).
Success on the EgoHOR sets grows
with the amount of EgoHOR (39.7, 42.6 and 47.2\% for none, 12.5\% and all of it).

\paragraph{EgoHOR matters most when public data is scarce.}
The less public data, the more EgoHOR helps the public sets ($+0.9$, $+3.3$ and $+5.4$ points with 100, 50 and
25\% of it; \cref{tab:ego}). With a quarter of the public data, it raises them from 43.1\% to 48.5\%, slightly
above training on twice the public data without it (47.2\%): references reconstructed from video can partly stand
in for public data.

\subsection{Ablations}
\label{sec:exp_ablation}

Each row of \cref{tab:ablation} changes only the components listed for it in \supp{tab:abl_rows} and keeps
everything else of Ours. Five rows remove a whole group of components; two split the object and scene inputs into
the world model's part and the policy's part. Every whole group matters: replacing one costs 6.0--17.8
zero-shot points (Ours varies by 3.9 across seeds) and 5.1--13.9 on the training clips; only replacing the world
model's object code and scene inputs with a per-object ID embedding leaves tracking success unchanged (below).

\begin{table}[t]
  \centering
  \footnotesize
  \setlength{\tabcolsep}{1.5pt}
  \definecolor{abldn}{rgb}{0.78,0.12,0.10}
  \newcommand{\dn}[1]{\textcolor{abldn}{\tiny$-$#1}}
  \newcommand{\up}[1]{\textcolor{gray}{\tiny$+$#1}}
  \caption{\textbf{Ablations}. SR$_{\text{MT}}$ (\%) and normalized return; WM: world model, $\pi$:
  policy; change relative to Ours in red (drop) or gray (gain); bold: best per column. Each row replaces some of our
  components (\supp{tab:abl_rows}, with per-set results).}
  \label{tab:ablation}
  \begin{tabular}{@{}l r@{\hspace{1pt}}l r @{\hspace{6pt}} r@{\hspace{1pt}}l r@{}}
    \toprule
    & \multicolumn{3}{c}{Training clips} & \multicolumn{3}{c}{Zero-shot} \\
    \cmidrule(lr){2-4} \cmidrule(l){5-7}
    & SR$_{\text{MT}}$ & & Return & SR$_{\text{MT}}$ & & Return \\
    \midrule
    Ours & \textbf{69.4} & & \textbf{0.688} & 51.0 & & \textbf{0.614} \\
    \midrule
    \multicolumn{7}{@{}l}{\textit{Ours without}} \\
    \quad contact geometry $\phi$ (WM)     & 55.5 & \dn{13.9} & 0.630 & 37.0 & \dn{14.0} & 0.552 \\
    \quad object, scene inputs (WM, $\pi$) & 60.8 & \dn{8.6}  & 0.657 & 33.2 & \dn{17.8} & 0.558 \\
    \quad object, scene inputs (WM)        & 68.8 & \dn{0.6}  & 0.682 & \textbf{51.9} & \up{0.9} & 0.613 \\
    \quad object inputs ($\pi$)            & 61.7 & \dn{7.7}  & 0.657 & 34.7 & \dn{16.3} & 0.561 \\
    \quad two-hand structure               & 63.3 & \dn{6.1}  & 0.668 & 43.2 & \dn{7.8}  & 0.586 \\
    \quad reward, training design          & 64.3 & \dn{5.1}  & 0.656 & 45.0 & \dn{6.0}  & 0.594 \\
    \quad exploration design               & 63.3 & \dn{6.1}  & 0.672 & 38.9 & \dn{12.1} & 0.580 \\
    \bottomrule
  \end{tabular}
\end{table}

\paragraph{Contact geometry and object inputs.}
Without the contact geometry, zero-shot success drops by 14.0 points and success on the training clips by
13.9: the world model's object position error after 24 steps nearly doubles on training and zero-shot clips alike
(\supp{tab:wm_err}).
The object inputs matter on the policy side: replacing its point-cloud encoding, contact geometry $\phi^h$ and local probes by
FDR's flat point cloud costs 16.3 points of zero-shot success but only 7.7 on the training clips, as expected if the
policy then partly memorizes objects and generalizes less well across them. In the world model, replacing the object code, scene distances and
hand-relative frames by a per-object ID embedding, support-surface parameters and the world frame changes
zero-shot success by only $+0.9$ points, although its object position error on unseen objects grows by 45\%
(\supp{tab:wm_err}): beyond the contact geometry, object information matters for tracking success in the policy, not in
the world model.

\paragraph{Two-hand structure, reward, and exploration.}
Replacing our two-hand structure (the mirrored module, the critic's inputs of the two hands and the allocation of
environments over single-hand and bimanual clips; \supp{sec:supp_critic}) costs 7.8 points zero-shot. Removing the squared kernels,
the rotation term, weight EMA, the real-return critic targets and modified learning hyperparameters costs 6.0, and
replacing our exploration design (noise blocks, return filter and finger
low-pass) by SuperMimic's uncorrelated Gaussian noise ($\sigma=0.1$) costs 12.1, with larger, temporally
correlated noise driving most of the gain (\supp{sec:supp_ablation}).

\subsection{Transfer to a second robot hand}
\label{sec:exp_sharpa}

To test whether Dexify depends on the XHAND or on elevated (4$\times$) finger torques, we train the same method
unchanged on the Sharpa Wave (22 finger joints per hand vs.\ 12 on the XHAND), using the same human clips, retargeting
pipeline, evaluation sets and hyperparameters; only hand-specific constants differ (\supp{sec:supp_sharpa}).

\begin{table}[t]
  \centering
  \small
  \setlength{\tabcolsep}{4pt}
  \caption{\textbf{Transfer to the Sharpa Wave}: SR$_{\text{MT}}$ (\%) on the training clips, mean over the seven
  zero-shot sets, and training GPU-hours on an RTX 4090 (--: not applicable or not timed on RTX 4090; full per-set
  results and ablations in \supp{tab:sharpa}). $^{\dagger}$Test-time optimization of the zero-shot clips.}
  \label{tab:exp_sharpa}
  \begin{tabular}{@{}ll rrr@{}}
    \toprule
    Torque & Method & Train & Zero-shot & GPU-h \\
    \midrule
    4$\times$ & XHAND, Ours                          & 69.4 & 51.0 & 15.3 \\
              & Sharpa, Ours                         & \textbf{80.7} & \textbf{71.7} & -- \\
    \midrule
    1$\times$ & XHAND, Ours                          & 55.2 & 31.0 & 15.5 \\
              & Sharpa, Ours                         & \textbf{81.3} & \textbf{70.3} & 17.0 \\
              & Sharpa, SPIDER~\cite{pan2025spider}  & --   & 52.0 & 150$^{\dagger}$ \\
              & Sharpa, SuperMimic~\cite{liao2026supermimic} & 23.8 & 16.1 & -- \\
              & Sharpa, FDR~\cite{lee2026fdr}        & 26.9 & 17.0 & $\approx$40 \\
    \bottomrule
  \end{tabular}
\end{table}

\begin{figure*}[t]
  \centering
  \includegraphics[width=\textwidth]{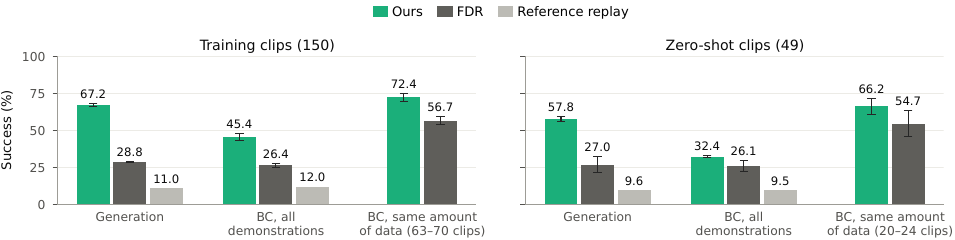}
  \caption{\textbf{Downstream imitation learning from converted trajectories.} Generation: share of generation
  perturbations executed successfully. BC: closed-loop test success of policies trained on all demonstrations, or on the
  same number of demonstrations per clip for source clips where both Dexify and FDR produced at least 10 demonstrations
  (63--70 of 150 training and 20--24 of 49 zero-shot clips). Ours and FDR: means over three generator seeds; error bars:
  standard deviation across seeds.}
  \label{fig:bc}
\end{figure*}

As shown in \cref{tab:exp_sharpa}, at 4$\times$ torque the Sharpa already reaches 71.7\% zero-shot and 80.7\%
training-clip success (vs.\ 51.0\% and 69.4\% on the XHAND), presumably owing to its additional finger joints and
kinematics closer to the human hand. More importantly, lowering finger torques from 4$\times$ to the real limits
(1$\times$) drops XHAND zero-shot success by 20.0 points (to 31.0\%), whereas the Sharpa loses only 1.4 points (to
70.3\%), presumably because its proximal finger joints are stronger (1.9--3.3 vs.\ 1.1\,N$\cdot$m). At 1$\times$ on the
Sharpa, our policy outperforms per-clip optimization SPIDER by 18.3 points (70.3 \vs 52.0\%, at roughly a ninth of its
150 GPU-hours) and model-free FDR by 53.3 points (17.0\%, trained for 600M steps). Ablations on the Sharpa at 1$\times$
confirm the same component ranking as on the XHAND (\supp{sec:supp_sharpa_abl}).

\subsection{Downstream imitation learning}
\label{sec:exp_bc}

To evaluate whether trajectories converted by Dexify serve as effective robot training data, we follow the
DexMimicGen~\cite{jiang2025dexmimicgen} protocol (\supp{sec:supp_bc}). On 150 training and 49 zero-shot source clips,
we perturb each clip's initial object pose (up to \qty{5}{cm} horizontally and \ang{20} yaw) to create 100 generation,
5 validation and 20 test perturbations per clip. Three generators execute the same generation perturbations at 4$\times$
torque: our policy, FDR (three generator seeds each, closed loop) and reference replay (open loop). Successful rollouts
become demonstrations of commanded PD targets and train identical closed-loop MLP behavior-cloning (BC) policies that
observe robot and object states, object surface points and target object poses, without human-hand inputs.

\Cref{fig:bc} summarizes the results. Dexify successfully executes more than twice as many generation perturbations as
FDR on both training and zero-shot clips, and about six times as many as reference replay. BC policies trained on
Dexify's demonstrations outperform those trained on FDR and reference replay both when using all generated
demonstrations and when matched to the same number of demonstrations per clip (paired Wilcoxon signed-rank test over
source clips, $p\le0.015$ for all comparisons), showing that the advantage is not only one of demonstration quantity.
Policies trained on our demonstrations also produce smoother finger commands (about \qty{25}{mrad} per step
vs.\ 63--\qty{73}{mrad} for FDR). Settings and per-seed zero-shot leads are in \supp{sec:supp_bc}.

\begin{figure}[t]
  \centering
  \includegraphics[width=\linewidth]{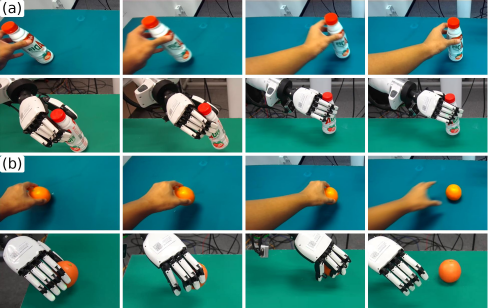}
  \caption{\textbf{Real-robot replay.} (a)~Milk bottle, (b)~orange. Top: the human recording, mirrored to
  match the robot's viewpoint; bottom: the retargeted motion replayed open loop on a Franka FR3 arm with an XHAND.}
  \label{fig:realrobot}
\end{figure}

\begin{figure*}[t]
  \centering
  \includegraphics[width=\textwidth]{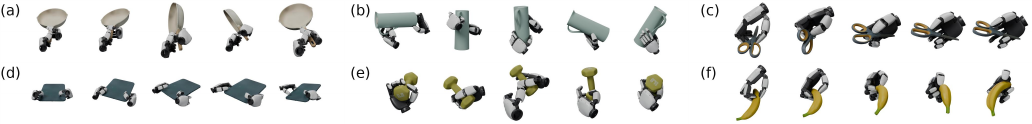}
  \caption{\textbf{Zero-shot tracking on public datasets}, five moments per clip: (a)~pan and (d)~cutting
  board, OakInk2-ZS; (b)~thermos and (e)~dumbbell, HOT3D-ZS; (c)~scissors and (f)~banana, one hand, DexYCB-ZS.
  Hands over \qty{30}{cm} from the object are not drawn.}
  \label{fig:gallery}
\end{figure*}

\subsection{Real-robot replay}
\label{sec:exp_real}

To test whether our motions transfer to hardware, we replay two retargeted motions on a real robot
(\cref{fig:realrobot}). We recorded a person picking up, carrying and placing a milk bottle and an orange,
reconstructed both recordings with our pipeline and tracked them in simulation with
Dexify trained at the real finger torque limits (1$\times$); neither object is in its training data. The tracked wrist trajectory is mapped to the joints of a Franka FR3 arm by inverse kinematics, and
the arm and an XHAND replay it with the finger targets.
In both clips the robot reproduces the recorded manipulation, suggesting that
motions retargeted in simulation can transfer to hardware.

\subsection{Qualitative results and failure modes}
\label{sec:exp_qual}

\Cref{fig:gallery} shows single-hand and bimanual zero-shot clips.
Most failures are not near misses: in 56\% of the failed zero-shot rollouts the mean object position error
exceeds \qty{10}{cm}, and only 7\% fail on the hand criteria alone.

\section{Conclusion}
\label{sec:conclusion}

Dexify retargets human hand--object interactions to dexterous robot hands with one policy, trained through a
differentiable world model. Both networks see the object through its geometry relative to the hand:
the world model sees the hand--object contact geometry, and the policy encodes the object relative to each hand. It converts 69.4\% of its training clips
and 51.0\% of zero-shot clips into physically feasible trajectories in simulation, \vs 31.1\% and 18.0\% for FDR at over twice the compute. It also converts as many zero-shot
clips as SPIDER, which optimizes each clip, at about a seventh of its compute. Its success grows with the
training data; EgoHOR raises it on its held-out categories and, when public data is scarce, on the public
sets. The same method transfers unchanged to a second hand (Sharpa Wave) at its real torque limits (70.3\%
zero-shot), and the converted trajectories outperform baselines for downstream imitation learning.

\paragraph{Limitations.}
Each clip has one rigid object, the scene is reduced to the surfaces it rests on, and a floating wrist stands in
for the robot arm. EgoHOR has no ground truth, and its object masses are estimated, not measured. On hardware we only
replay motions open loop; closed-loop downstream policy training is evaluated in simulation (\cref{sec:exp_bc}). Many
zero-shot clips still fail, mostly by losing the object.

\section*{LLM Usage Statement}

We used an LLM-based coding and writing assistant (Claude, Anthropic) in this work.
\emph{Code:} it wrote and modified parts of the training, evaluation and data-processing code, the
reimplementation of the FDR baseline in our simulator, the analysis scripts and the plotting code of the figures.
\emph{Experiments:} it launched and monitored training and evaluation jobs and summarized their logs into
the reported tables.
\emph{Writing:} it drafted and edited parts of the text and the tables.
The authors made the research decisions, checked the code, results and text, and take full
responsibility for the content.
Separately, the reconstruction pipeline of \cref{sec:egorecon} uses a vision--language model (GPT, OpenAI) to
select interaction segments, give the segmentation prompt and estimate the object's mass.
The reconstruction pipeline's code and the code base this work builds on were also written with LLM coding
assistants.

{
    \small
    \bibliographystyle{ieeenat_fullname}
    \bibliography{main}
}

\clearpage
\setcounter{page}{1}
\maketitlesupplementary
\raggedbottom

\appendix

\renewcommand{\topfraction}{0.9}\renewcommand{\dbltopfraction}{0.9}\renewcommand{\textfraction}{0.07}
\renewcommand{\floatpagefraction}{0.6}\renewcommand{\dblfloatpagefraction}{0.6}
\setcounter{topnumber}{4}\setcounter{dbltopnumber}{3}\setcounter{totalnumber}{6}
\makeatletter
\setlength{\@fptop}{0pt}\setlength{\@fpsep}{12pt plus 2pt}\setlength{\@fpbot}{0pt plus 1fil}
\setlength{\@dblfptop}{0pt}\setlength{\@dblfpsep}{12pt plus 2pt}\setlength{\@dblfpbot}{0pt plus 1fil}
\makeatother

\makeatletter
\newcommand{\suppbibnum}[2]{\expandafter\def\csname sbn@#1\endcsname{#2}}
\newcommand{\citesupp}[1]{[\def\sc@sep{}\@for\sc@k:=#1\do{\sc@sep\hyperlink{sref:\sc@k}{\csname sbn@\sc@k\endcsname}\def\sc@sep{, }}]}
\makeatother
\suppbibnum{genrobot2026genhumanego}{S1}
\suppbibnum{jiang2025dexmimicgen}{S2}
\suppbibnum{lee2026fdr}{S3}
\suppbibnum{liao2026supermimic}{S4}
\suppbibnum{lightwheel2026egosuite}{S5}
\suppbibnum{mandi2025dexmachina}{S6}
\suppbibnum{paliwal2026doasido}{S7}
\suppbibnum{pan2025spider}{S8}
\suppbibnum{wei2022coacd}{S9}
\suppbibnum{zhu2026chord}{S10}

\newsavebox{\fitcolbox}
\newcommand{\fitcol}[1]{\sbox{\fitcolbox}{#1}\ifdim\wd\fitcolbox>\linewidth
  \resizebox{\linewidth}{!}{\usebox{\fitcolbox}}\else\usebox{\fitcolbox}\fi}
\definecolor{suppdn}{rgb}{0.78,0.12,0.10}
\newcommand{\sdn}[1]{\textcolor{suppdn}{\tiny$-$#1}}
\newcommand{\sgn}[1]{\textcolor{gray}{\tiny$+$#1}}
\newcommand{\serrok}[1]{\colorbox{gray!15}{\tiny\textcolor{darkgray}{$+$#1\%}}}
\newcommand{\serrhi}[1]{\colorbox{suppdn!15}{\tiny\textcolor{suppdn}{$+$#1\%}}}

\etocsettagdepth{mtmain}{none}
\etocsettagdepth{mtsupp}{subsection}
\tableofcontents
\etocdepthtag.toc{mtsupp}
\vspace{\baselineskip}

This supplement first details the downstream imitation-learning evaluation of \cref{sec:exp_bc}
(\cref{sec:supp_grp_quality}). It then gives the details of the data (\cref{sec:supp_data}) and the
method (\cref{sec:supp_grp_method}), and the details and full results of the experiments (\cref{sec:supp_grp_experiments}),
including the results at the XHAND's real finger torque limits (1$\times$; the main results use four times these
limits, 4$\times$). Finally, it gives the full setup, per-set results and ablations for the second robot hand of
\cref{sec:exp_sharpa} (\cref{sec:supp_sharpa}).

\section{Downstream Imitation Learning}
\label{sec:supp_grp_quality}\label{sec:supp_bc}

This section details the downstream imitation-learning evaluation summarized in \cref{sec:exp_bc}. Following the
protocol of DexMimicGen~\citesupp{jiang2025dexmimicgen},
the object's initial pose in a set of source clips is perturbed, the executions that succeed become demonstrations, and a
behavior-cloning (BC) policy is trained on them. Three generators execute the same perturbations: our policy, FDR and
reference replay (\cref{sec:exp_baselines}). Only the generator differs; the BC policies, their training and the
selection of their settings are the same for all. Our policy generates more than twice as many demonstrations as FDR,
and policies trained on them outperform the others, also with the same amount of data (\cref{fig:bc}).

\subsection{Demonstration generation}
\label{sec:supp_bc_gen}

Each demonstration starts from a human source clip. We choose source clips by their motion alone, not by whether a
tracker succeeds on them: the object is clearly moved (10 to \qty{30}{cm}), and the hands reach it from a distance,
which leaves room to redirect the approach; each object contributes at most one clip. We use 150 training clips, on
which both our policy and FDR were trained, and 49 zero-shot clips, on which neither was.

Each source clip is then multiplied by perturbing the object's initial pose. A perturbation moves this pose horizontally by up to
\qty{5}{cm} and rotates it by up to \ang{20} about the vertical axis through its initial position. The same rigid
transform is applied to the object's whole trajectory and its support surfaces. The hands start where the human's did
and move gradually onto the transformed path, which they reach shortly before the first contact; from then on, hands
and object follow the source motion under the same transform. Each source clip has 100 generation perturbations for
the generators, and 5 validation and 20 test perturbations for the BC policies.

The generators turn the generation perturbations into robot demonstrations. Our policy and FDR are the final policies
of the main comparison (three seeds each; ours after 40k iterations, about 273M simulator steps, FDR after 600M steps)
and run in closed loop; reference replay runs open loop on the transformed kinematic reference, as in DexMimicGen. All
three execute every generation perturbation with four times the XHAND's finger torque limits (4$\times$), the setting of
the main results. Every rollout that meets the object criterion of SR$_{\text{MT}}$ (mean object position and rotation
errors within \qty{3}{cm} and \ang{30})
becomes a demonstration of its commanded PD targets.

\subsection{Behavior cloning}
\label{sec:supp_bc_bc}

The demonstrations of each generator then train a BC policy, an MLP that observes the state of the hands and the object, the object's
upcoming target poses and its surface points, but not the human hand, and predicts a chunk of PD-target increments.
The source clips are different tasks, so the object's target poses tell one policy which task to perform.
\Cref{tab:bc_hparams} lists its settings. We choose one network size for all policies, by the success of the policies
trained on the demonstrations of the first generator seed on the validation perturbations.

We train the BC policy on two kinds of data. First, on all demonstrations that a
generator produced: this compares what the generators obtain from the same perturbations, so a generator that succeeds more
often also provides more data. Second, on the same amount of data: for every source clip on which our policy and FDR
each produced at least 10 demonstrations, both training sets take the same number of demonstrations of that clip.
This compares the generators at the same amount of data per clip, but only on the clips that both trackers handle: the harder
clips, which FDR rarely completes and our policy often does, are left out.

The trained policies are evaluated closed loop on the test perturbations, by the object criterion; the results
of our policy and FDR average three generator seeds, each with three BC seeds.

\begin{table}[!t]
  \centering
  \small
  \caption{\textbf{Behavior-cloning settings}, the same for all generators.}
  \label{tab:bc_hparams}
  \begin{tabular}{@{}ll@{}}
    \toprule
    Network & MLP, $3\times1024$, $4\times2048$ or $4\times4096$ (chosen) \\
    Action chunk & 16 PD-target increments, first 4 executed \\
    Training & 40k steps, batch 4096 \\
    Optimizer & AdamW, weight decay $10^{-4}$ \\
    Learning rate & $3\cdot10^{-4}$, 1k warm-up steps, cosine decay \\
    Checkpoint & lowest loss on 5\% held-out demonstrations \\
    \bottomrule
  \end{tabular}
\end{table}

\subsection{Results}
\label{sec:supp_bc_results}

Our policy turns more of the perturbations into demonstrations: it successfully executes more than twice as many generation perturbations as FDR, on
the training and on the zero-shot clips, and about six times as many as reference replay (\cref{fig:bc}). These demonstrations also make better training data: policies
trained on them outperform those trained on FDR's and on reference
replay's, with all demonstrations and, on the clips where both trackers produce enough, with the same amount of data
(\cref{fig:bc}); on these clips the advantage is therefore not only one of quantity. All these differences are
significant (paired Wilcoxon signed-rank test over source clips, $p\le0.015$).

On the zero-shot clips, the lead over FDR is smaller and varies with the generator seed: 11.6, 3.3 and 4.1 points for
the three seeds with all demonstrations, and 26.6, 8.7 and $-0.6$ points with the same amount of data. The policies trained on
our demonstrations also move more smoothly: their finger targets change by about \qty{25}{mrad} per step, against 63 to
\qty{73}{mrad} for those trained on FDR's (first generator seed).

Like DexMimicGen's, our evaluation stays within the source tasks: the test perturbations come from the same source clips as the
generation perturbations, so the results measure the converted trajectories as demonstrations for these tasks.

\section{Constructing Hand--Object References}
\label{sec:supp_data}

This section describes how the human references of \cref{sec:data} are obtained: EgoHOR clips are reconstructed from
egocentric video (\cref{sec:supp_recon}), the public datasets are adapted to our setting (\cref{sec:supp_public}), and
all clips then pass through the same kinematic retargeting and filtering (\cref{sec:supp_processing}).

\subsection{Reconstruction from egocentric video}
\label{sec:supp_recon}

\begin{figure*}[!t]
  \centering
  \includegraphics[width=0.9\textwidth]{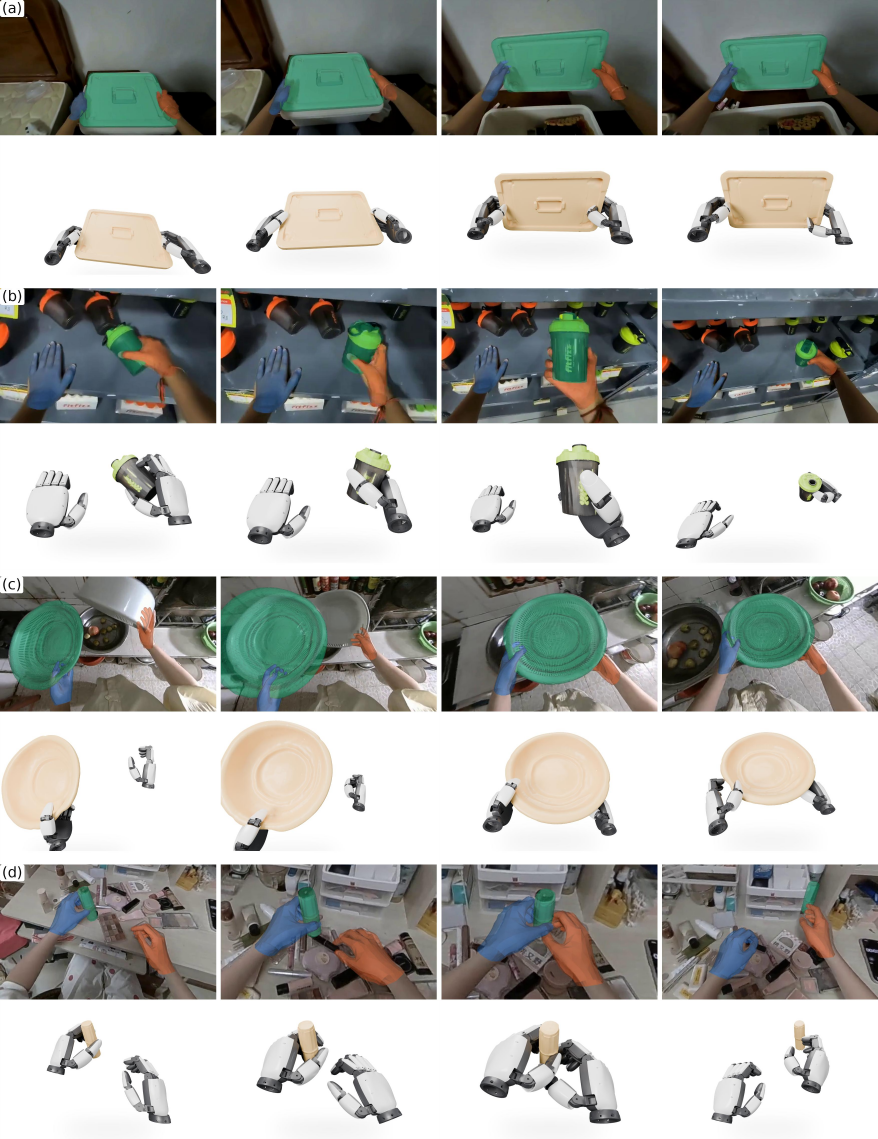}
  \caption{\textbf{Reconstructions from egocentric video and their zero-shot tracking.} Clips of held-out
  categories from EgoStd-ZS, (a)~box lid and (b)~shaker, and from GenHE-ZS, (c)~basket and (d)~bottle.
  For each clip, top: the video frame with the reconstructed hands (blue, orange) and object (green) overlaid;
  bottom: our policy tracking the clip zero-shot in simulation at the same moment, seen from the same headset
  camera. The lid, basket and bottle are drawn in a uniform color, as their reconstructed colors are close
  to those of the hands.}
  \label{fig:supp_recon}
\end{figure*}

\paragraph{Sources.} EgoStandard is part of EgoSuite-Open100K~\citesupp{lightwheel2026egosuite}: calibrated stereo video and
hand keypoints from a headset; a vision--language model selects the segments in which the hands interact with one
rigid object. Gen-HumanEgo~\citesupp{genrobot2026genhumanego} also records stereo video, from the forward pair of six fisheye
cameras, together with both hands' keypoints and task and subtask annotations; we rectify the stereo pair and process
it with the same pipeline as EgoStandard. Its segments are the annotated subtasks of picking up or putting down a rigid
object; small and non-rigid objects are excluded. For both sources, the vision--language model
provides the point prompt for segmentation and estimates the object's mass. \Cref{fig:supp_recon}
shows reconstructed clips from both sources and our policy tracking them zero-shot.

\paragraph{Object meshes and poses.} Reconstructed meshes are often open shells with holes. We use them as they are:
filling the holes to make solids collapses 76\% of them to their convex hulls, so that cups and baskets, for example,
become solid blocks that the hands can no longer hold as in the video.

With the mesh fixed, the object's 6D trajectory is initialized by robust rigid registration of
the tracked 3D points, then refined, coarse to fine, against the visible silhouette and depth with
penalties on translational and angular acceleration, keeping mesh, scale, camera and hands fixed. Where the
fine stage fails its silhouette, depth or motion checks, the coarse solution is kept. Pose fitting does not enforce
physical stability, so an object at rest often remains tilted away from a pose in which it could rest on its support,
by a median of \ang{10} in the clips we keep; clips in which it cannot rest at the start are discarded by the filters
(\cref{sec:supp_processing}).

\paragraph{Contact repair.} We call a frame untouched when the object moves while no MANO joint is within
\qty{2}{cm} of it; such stretches longer than \qty{0.25}{s} occur in 15\% of the EgoStandard and 32\% of the
Gen-HumanEgo training clips, against 0 to 8\% in the public datasets, which makes these clips hard to track. Most
untouched frames of rejected clips come after the last contact (the reconstructed object keeps drifting or
stops where no support surface is placed) or from depth drift while a hand is within \qty{5}{cm}.

We therefore repair the object's trajectory in the human reference and leave the hand unchanged, in three phases of
the clip. Before the first contact, the object stays at its initial pose. After the last contact, it stays where the
hand left it, and is set down on its support when it is just above it. In between, where it moves with no fingertip
nearby, it is shifted toward the nearest fingertip, unless no fingertip is within reach. All changes are blended in
over a few frames, and in training clips the shift is reduced where it would push the hand into the object.

Repaired clips are then checked again. A repaired clip must pass the filters (\cref{sec:supp_processing}) before retargeting and again after it. Repaired clips that still have
untouched stretches longer than \qty{0.25}{s} are discarded, and so are repaired training clips whose object
jumps by more than \qty{0.5}{cm} more per frame than in the original.

\paragraph{Counting distinct objects.}\label{sec:supp_count} The same object is reconstructed anew in every recording
or subtask in which it appears, so we count distinct objects by clustering the meshes that carry the same object noun
by their shape distance (\cref{sec:supp_zs}). The threshold, \qty{0.75}{cm}, separates repeated reconstructions of one
object (median distance about \qty{0.5}{cm}) from different objects with the same noun (about \qty{1}{cm}). With it,
the 8451 EgoStandard and 2984 Gen-HumanEgo reconstructions form 1862 and 990 objects; thresholds of 1 and
\qty{0.5}{cm} give 1222 to 3382 and 663 to 1731.

\paragraph{Retention.} \Cref{tab:supp_funnel} follows both sources from the candidates to the training clips. Most
losses occur at two steps: about half of the reconstructed segments fail the checks that the object's mask covers and
aligns with the object, and about half of the imported clips fail retargeting and its filters
(\cref{sec:supp_processing}). Repair recovers many of the rejected clips: about a third of the training clips are included only
after contact repair or support recovery, most of them clips that the filters had rejected; support recovery re-imports
clips rejected only because the resting object starts tilted or ends below its support, adds a support surface under
its final rest, and applies the same filters again. Between the filters and training, the clips of held-out categories go to the zero-shot sets, and
quality checks of the retargeted reference (palm jumps, penetration, object jumps, among others) remove further clips;
for Gen-HumanEgo, most of these are clips whose object still floats or penetrates its support.

\begin{table}[!t]
  \centering
  \footnotesize
  \setlength{\tabcolsep}{4pt}
  \caption{\textbf{EgoHOR retention}, clips per stage.
  $^{a}$EgoStandard: whole recordings, chosen by their text descriptions from a catalog of 295637; Gen-HumanEgo:
  annotated pick-up and put-down subtasks. Not counted: 26646 and 5452 candidates that were never processed
  because of processing failures.
  $^{b}$One clip per object, so a scene with two objects counts twice.
  $^{c}$Before repair: the training clips include clips that failed these filters and pass them after support recovery
  or contact repair, so the counts are not nested.}
  \label{tab:supp_funnel}
  \begin{tabular}{@{}lrr@{}}
    \toprule
    Stage & EgoStandard & Gen-HumanEgo \\
    \midrule
    Candidates processed$^{a}$          & 59755           & 31483 \\
    Segments reconstructed              & 28220    & 28418 \\
    Pass reconstruction checks          & 13457    & 11571 \\
    Clips imported$^{b}$                & 15412    & 11571 \\
    Pass retargeting and filters$^{c}$  & 8101     & 5610 \\
    \midrule
    Training clips                      & 8451     & 2984 \\
    \quad unchanged                     & 5894            & 1882 \\
    \quad support recovered             & 505                & 202 \\
    \quad contact repaired              & 2052            & 900 \\
    Zero-shot clips                     & 115                & 235 \\
    \bottomrule
  \end{tabular}
\end{table}

\subsection{Public datasets}
\label{sec:supp_public}

GRAB and HOT3D are used as provided; HOT3D contributes both single- and bimanual clips. The other public datasets need
adapting to our setting. Where each hand holds its own object (OakInk2, TACO), we split the interaction into one clip
per object and end it before the two objects touch, since tool--target contact is not simulated; in each such clip, the
other hand still follows its human motion, but without its object. From DexCanvas, which has many clips of few objects, we keep
at most 500 clips per category, and drop one object whose reference poses jump between frames.
HRDexDB records grasps of the same objects by a human hand and by robot hands; we use both, taking the Allegro and
Inspire hands' wrist and fingertips from their forward kinematics in place of MANO. From HOI4D we keep pick-and-place
tasks; its annotations are given in the moving camera's frame, and we express them in the world frame with the camera
trajectory and hold resting objects fixed.

\subsection{Kinematic retargeting and filtering}
\label{sec:supp_processing}

\paragraph{Retargeting.} We map MANO to the robot hand with FDR's kinematic retargeting, after expressing every
reference in a z-up world frame and resampling it to the \qty{60}{\hertz} control rate. Its inverse kinematics restarts
within some clips, which makes the wrist joints jump; we detect these jumps and correct them. For simulation, each object
is decomposed into convex parts for collision~\citesupp{wei2022coacd}, and given a signed distance grid for the contact
geometry $\phi$.

\paragraph{Support surfaces.}\label{sec:supp_scene} The scene $\mathcal{S}$ is a union of at most four support surfaces per clip, each a horizontal disk
\qty{1}{cm} thick whose top lies at the height where the object rests.
One is placed at the object's start pose and one at every later pose where the reference object rests
untouched; overlapping surfaces at nearly the same height are merged.
Disk radii are rounded up to \qtylist{8;15;25;40}{cm}.
Their signed distance, used for the scene geometry $\psi$ that the world model receives (\cref{eq:phi}), is computed
analytically and is differentiable.

\paragraph{Filtering.} We discard references whose recorded motion cannot be reproduced under physics or that would teach wrong
behavior: clips in which the object is unsupported at the start, ends below its support or moves without being
touched, clips shorter than \qty{1}{s}, and clips whose retargeting still produces palm jumps or wrist flips.

\section{Method}
\label{sec:supp_grp_method}

\Cref{sec:method} describes the method as far as needed to understand it; this section gives what is needed to
reproduce it: the simulated task (\cref{sec:supp_sim}), the full reward terms (\cref{sec:supp_reward}) and world-model
inputs (\cref{sec:supp_wm}), the two-hand structure (\cref{sec:supp_critic}), and all hyperparameters
(\cref{sec:supp_hparams}).

\subsection{Simulation}
\label{sec:supp_sim}

\paragraph{Finger torques.} The torque of each finger joint, the sum of its PD terms, is limited to four times the
XHAND's specified limits (1.1 or \qty{0.4}{N.m}, depending on the joint), denoted 4$\times$; 1$\times$ uses the
specified limits. At 1$\times$, the XHAND's fingers often cannot hold the object: from 4$\times$ to 1$\times$, our
policy's zero-shot success falls by 20.0 points on the XHAND but by only 1.4 points on a second hand whose proximal
finger joints are stronger (\cref{sec:supp_sharpa_xhand}). The main results therefore use 4$\times$.

\paragraph{Wrist.} In simulation, the hand moves on a floating wrist with six actuated joints. In practice, such a hand
is mounted on a robot arm, so we limit the wrist to what an arm could provide: every hand link's weight is compensated,
the six wrist joints' PD gains give a critically damped \qty{5}{\hertz} bandwidth, and the wrist force and
torque are limited to \qty{100}{N} and \qty{12}{N.m}, comparable to the wrist joints of a collaborative
arm. The wrist PD targets are low-pass filtered at \qty{5}{\hertz} and limited to \qty{3}{m/s} and \qty{20}{m/s^2} in translation, and
\qty{6}{rad/s} and \qty{50}{rad/s^2} in rotation by a second-order limiter that decelerates before reaching a bound, both in
simulation and in imagined rollouts; about 97\% of the human clips stay within these bounds.

\paragraph{Single-hand clips.} In a single-hand clip the other hand is idle: it is parked away from the workspace
and its actions are ignored, so single-hand and bimanual clips run in the same bimanual simulation.

\paragraph{Contacts.} We use MuJoCo's elliptic friction cone with an impedance ratio of 10; the MuJoCo documentation
recommends elliptic cones with a large impedance ratio to reduce slip, which is especially problematic in manipulation.

\subsection{Reward terms}
\label{sec:supp_reward}

As in \cref{eq:reward}, the reward of each hand is
\begin{equation}
  r_t = r^{\text{FDR}}_t + w_{\text{rot}}r_{\text{rot}} + w_{\text{vel}}r_{\text{vel}}
        + w_{\Delta\text{w}}r_{\Delta\text{w}} + w_{\text{jerk}}r_{\text{jerk}},
  \label{eq:supp_reward}
\end{equation}
and the step reward is its mean over the active hands. As in \cref{sec:training}, the base reward $r^{\text{FDR}}$
adapts FDR's reward with squared kernels in $r_{\text{pcd}}$ and $r_{\text{tip}}$, an action-rate penalty (the
$w_{\Delta}$ term of $r_{\text{reg}}$) and no termination penalty,
$r^{\text{FDR}}=w_{\text{pcd}}r_{\text{pcd}}+w_{\text{int}}r_{\text{int}}+r_{\text{imit}}+r_{\text{reg}}$, with
\begin{equation}
  \begin{aligned}
    r_{\text{pcd}} &= \exp\!\big(-\tfrac{1}{3}\textstyle\sum_{m\in\mathcal{K}_t} e_m^2/\sigma_{\text{pcd}}^2\big),\\
    r_{\text{int}} &= \tfrac{1}{5}\textstyle\sum_{f}\exp\!\big(-\max(0,\,d_f-\hat{d}_f)/\sigma_{\text{int}}\big),\\
    r_{\text{imit}} &= w_{\text{wp}}r_{\text{wp}} + w_{\text{wr}}r_{\text{wr}} + w_{\text{tip}}r_{\text{tip}} + w_{\text{bc}}r_{\text{bc}},\\
    r_{\text{wp}} &= \exp\!\big(-10\,\lVert x_{\text{w}}-\hat{x}_{\text{w}}\rVert\big),\\
    r_{\text{wr}} &= \exp\!\big(-3\,\angle(R_{\text{w}},\hat{R}_{\text{w}})\big),\\
    r_{\text{tip}} &= \tfrac{1}{5}\textstyle\sum_{f}\exp\!\big(-\lVert x_f-\hat{x}_f\rVert^2/\sigma_{\text{tip}}^2\big),\\
    r_{\text{bc}} &= \tfrac{1}{J}\textstyle\sum_{j}\exp\!\big(-\kappa\,(q^{\text{pd}}_j-\hat{q}_j)^2/\Delta_j\big),\\
    r_{\text{reg}} &= w_a\,\overline{a_t^2} + w_{\Delta}\lVert a_t-a_{t-1}\rVert^2.
  \end{aligned}
  \label{eq:terms}
\end{equation}
The other terms of \cref{eq:supp_reward} are the object rotation term $r_{\text{rot}}=\exp(-\angle(R_t,\hat{R}_t))$
(\cref{sec:training}) and the smoothness terms
\begin{equation}
  \begin{aligned}
    r_{\text{vel}} &= \textstyle\sum_{u\in\{v,\,\omega,\,v^{\text{w}},\,\omega^{\text{w}}\}}
      \exp\!\big(-\lVert u_t-\hat{u}_t\rVert^2/\sigma_u^2\big),\\
    r_{\Delta\text{w}} &= \lVert a^{\text{w}}_t-a^{\text{w}}_{t-1}\rVert^2 ,\\
    r_{\text{jerk}} &= \lVert a_t-2a_{t-1}+a_{t-2}\rVert^2 ,
  \end{aligned}
  \label{eq:rsmooth}
\end{equation}
where $v_t,\omega_t$ and $v^{\text{w}}_t,\omega^{\text{w}}_t$ are the linear and angular velocities of the object
and of the wrist, $a_t$ all of the hand's actions after clipping and low-pass filtering and $a^{\text{w}}_t$ its six
wrist actions; $r_{\text{jerk}}$ is
zero for actions that change at a constant rate and grows with the frequency of their changes.
Here $e_m=\lVert(R_tp_m+p_t)-(\hat{R}_tp_m+\hat{p}_t)\rVert$ is the error of the object's surface point $p_m$ (128
points, rescaled to fit a \qty{5.8}{cm} radius) and $\mathcal{K}_t$ holds the three largest errors; $d_f$ and $\hat{d}_f$
are the signed distances of the robot and the MANO fingertip $f$ to the object surface (so $r_{\text{int}}$ penalizes
only fingertips farther from the surface than the human's); $x_{\text{w}},R_{\text{w}}$ and $\hat{x}_{\text{w}},\hat{R}_{\text{w}}$ are the poses of
the palm and of the MANO wrist, $x_f$ and $\hat{x}_f$ the robot and MANO fingertips, $q^{\text{pd}}$ the commanded and
$\hat{q}$ the retargeted joint targets, and $\Delta_j$ the joint ranges. All terms are multiplied by the control period;
\cref{tab:hparams_task} lists the weights and constants.

\subsection{World-model inputs and state update}
\label{sec:supp_wm}

As in \cref{eq:wmstep}, the world model predicts increments of the state from hand-relative inputs,
\begin{equation}
  f_\omega(\hat{s}_t,a_t)=\hat{s}_t\oplus\delta_\omega\big(u_t,\,a_t,\,\phi_t,\,\psi_t,\,z_o\big),
\end{equation}
where $u_t$ are features of the state $\hat{s}_t$, $a_t$ the actions of both hands, $\phi_t$ and $\psi_t$ the contact
and scene geometry (\cref{eq:phi}), $z_o$ a code of the object, and $\oplus$ applies the predicted increments
$\delta_\omega$ to the state. We detail the inputs $u_t$ and $a_t$, the object code and the state update below; the point sets
of $\phi$ and $\psi$ and their clipping ranges are listed in \cref{tab:hparams_learn}.

\paragraph{State and action inputs.} Neither the inputs nor the outputs depend on where in the workspace an interaction
happens, so what the model learns at one position applies at any other. For each hand, the state features $u_t$ hold its wrist height,
the gravity direction in its wrist frame instead of the world-frame wrist orientation, its wrist velocities in its own
frame, the object's pose and velocities in its wrist frame, and its joint positions and velocities without the wrist's
two horizontal positions. The action enters as the PD target minus the current joint positions.

\paragraph{Object code.} The object code summarizes the object's shape and mass properties,
\begin{equation}
  z_o = g\Big(\big[\max_{m} h_o(p^o_m),\;\mu_o\big]\Big),
  \label{eq:objenc}
\end{equation}
where $h_o$ is a per-point MLP on surface points $p^o_m$ in the object's own frame at true scale, the maximum is taken
per feature, $\mu_o$ holds the logarithm of the object's mass, its center of mass and its inertia per unit mass, and $g$
is an MLP producing the code.

\paragraph{State update.} The increments are expressed relative to the hand, like the inputs: $\oplus$ adds the joint
increments, composes the object orientation as $R_t\exp(\Delta\theta_t)$, and adds the object's position and velocity
increments after rotating them from the right hand's wrist frame to the world.

\subsection{Two-hand structure: critic inputs and environment allocation}
\label{sec:supp_critic}

\paragraph{Critic inputs.} The critic's features for hand $h$ (\cref{sec:training}) are
\begin{equation}
  \zeta^h_t=\big(u^h,\;e^h_t,\;\hat{x}^h_{>t},\;d^h_t,\;[p_m,\Delta p_m]_m,\;\phi^h_{\text{tip}}\big),
  \label{eq:critic}
\end{equation}
where $u^h$ is hand $h$'s part of the world model's state features $u$ (\cref{eq:wmstep}), $e^h_t$ the
errors of the object pose and the fingertips to the reference in the wrist frame, $\hat{x}^h_{>t}$ the
upcoming reference frames (\cref{eq:obs}), $d^h_t$ features of the support surfaces, $[p_m,\Delta p_m]_m$
all of the object's points in the wrist frame and their displacements to the next reference pose, as in the
policy's point-cloud encoding but as one flat vector, and $\phi^h_{\text{tip}}$ the contact geometry of
\cref{eq:phi} at the palm and the fingertips. The critic receives both hands' features side by side, with an idle hand's
features set to zero. The point cloud is included only for the right hand: the left hand's would describe the same
object in another frame, and the object's pose relative to the left wrist is already among the left hand's features.

\paragraph{Environment allocation.}\label{sec:supp_alloc} Each simulated environment holds one object. Half of the
environments run bimanual and half single-hand clips, and within each half an object receives environments in
proportion to the square root of its frames. Sources with only one or a few clips per object receive a fixed share
instead, so that their thousands of objects do not dominate the allocation. Objects are redrawn every 1000 iterations, so that training covers more objects than there are
environments. Each clip is cut into \qty{1}{s} bins, and an episode starts at a random frame of a bin chosen half of the
time uniformly and half of the time by SuperMimic's prioritized reset, which favors bins where recent episodes earned low
returns (\cref{tab:hparams_task}).

\subsection{Hyperparameters}
\label{sec:supp_hparams}

\Cref{tab:hparams_task} lists the simulation, environment and reward settings of the final configuration, and \cref{tab:hparams_learn} its learning hyperparameters.
We keep an exponential moving average of the policy weights (Weight EMA in \cref{tab:hparams_learn}) and
evaluate it. The local probes $\chi^h$ of the palm and the fingertips share one MLP. The action noise of a block
with level $\sigma$ (Action-noise blocks in \cref{tab:hparams_learn}) is $\sigma\varepsilon_t$, where
$\varepsilon_t=\rho\,\varepsilon_{t-1}+\sqrt{1-\rho^2}\,\xi_t$ with $\xi_t\sim\mathcal{N}(0,I)$ per control step, drawn
anew at each episode start: each step's noise has the same variance as independent noise, but with $\rho=0.8$ it
changes smoothly, with a correlation time of about \qty{75}{ms}.

\begin{table*}[!t]
  \centering
  \small
  \caption{Simulation, environment and reward settings of the final configuration.}
  \label{tab:hparams_task}
  \begin{tabular}{@{}p{0.33\textwidth}p{0.64\textwidth}@{}}
    \toprule
    \multicolumn{2}{@{}l}{\textbf{Simulation}} \\
    Simulator & MuJoCo-Warp, condim 3, elliptic cone, impratio 10 \\
    Contact softness & MuJoCo's defaults: solref $(0.02, 1)$, solimp $(0.9, 0.95, 0.001, 0.5, 2)$ \\
    Finger torque limits & 4$\times$ the XHAND's (1$\times$ where stated) \\
    Wrist & gravity-compensated; PD \qty{5}{\hertz}, critically damped; \qty{100}{N} / \qty{12}{N.m} \\
    Wrist-target limits & \qty{3}{m/s}, \qty{20}{m/s^2}, \qty{6}{rad/s}, \qty{50}{rad/s^2} \\
    Solver / line-search iterations & 100 / 50 \\
    Physics / control rate & \qty{240}{\hertz} / \qty{60}{\hertz} \\
    PD-target low-pass, wrist / fingers & \qty{5}{\hertz} / \qty{8}{\hertz} \\
    Parallel environments & 4096 \\
    \midrule
    \multicolumn{2}{@{}l}{\textbf{Environment allocation and resets}} \\
    Bimanual / single-hand clips & 50\% / 50\% of the environments \\
    Environments per object & $\propto\sqrt{\text{frames}}$ \\
    Fixed source shares & EgoHOR 50\% of bimanual, HOI4D 12\% of single-hand \\
    Objects redrawn & every 1000 iterations \\
    Reset bins & \qty{1}{s}; 50\% uniform, 50\% prioritized \\
    \midrule
    \textbf{Reward} (\cref{eq:reward,eq:terms,eq:rsmooth}) & \emph{weight; constants} \\
    $r_{\text{pcd}}$ & 1.0; $\sigma_{\text{pcd}}=\qty{5.48}{cm}$, worst 3 of 128 points \\
    $r_{\text{rot}}$ & 2.0 \\
    $r_{\text{vel}}$ & 1.0; $\sigma_v$, $\sigma_\omega$ = \qty{0.2}{m/s}, \qty{1.5}{rad/s} (object) and \qty{0.15}{m/s}, \qty{1}{rad/s} (wrist) \\
    $r_{\text{int}}$ & 0.3; $\sigma_{\text{int}}=\qty{1}{cm}$ \\
    $r_{\text{wp}}$ / $r_{\text{wr}}$ & 0.15 / 0.15 \\
    $r_{\text{tip}}$ & 0.3; $\sigma_{\text{tip}}=\qty{3.87}{cm}$ \\
    $r_{\text{bc}}$ & 0.3; $\kappa=250$ \\
    $w_a$ / $w_\Delta$ / $w_{\Delta\text{w}}$ / $w_{\text{jerk}}$ & $-0.01$ / $-0.1$ / $-10$ / $-3$ \\
    Termination penalty & 0 \\
    \bottomrule
  \end{tabular}
\end{table*}

\begin{table*}[!t]
  \centering
  \small
  \caption{Learning hyperparameters of the final configuration, shared by all runs unless an ablation
  changes them.}
  \label{tab:hparams_learn}
  \begin{tabular}{@{}p{0.33\textwidth}p{0.64\textwidth}@{}}
    \toprule
    \multicolumn{2}{@{}l}{\textbf{World model}} \\
    Network & MLP, 3 hidden layers of 2048, SiLU, LayerNorm \\
    Fit window $k$ / batch / batches per iteration & 8 steps / 2048 windows / 16 \\
    Contact geometry $\phi$ & palm, 5 fingertips and 17 link points per hand; SDF clamped to $[-5,10]$\,cm \\
    Scene geometry $\psi$ & 16 object surface points, palm and 5 fingertips per hand; SDF clamped to $[-2,30]$\,cm \\
    Object code & 16-d, from a PointNet on 128 surface points, the log mass, center of mass and inertia per unit mass \\
    Replay buffer & 1000 vectorized steps $\times$ 4096 environments \\
    World-model-only warm-up & 200 iterations \\
    \midrule
    \multicolumn{2}{@{}l}{\textbf{Policy}} \\
    Goal encoder / decoder & MLP 2048-1024-512-256; goal latent 32 \\
    Observation history / goal frames & 10 / 10 \\
    Point-cloud encoder & PointNet on the first 64 of the 128 points, per-point MLP 64-128-128, max pooling \\
    Local probes $\chi^h$ & $3{\times}3{\times}3$ grid, \qty{1.5}{cm} spacing, around the palm and each fingertip; shared MLP \\
    Other-hand summary & 64 \\
    Imagination horizon $H$ / imagined starts & 24 / 4096 \\
    Weight EMA & 0.9995 \\
    \midrule
    \multicolumn{2}{@{}l}{\textbf{Critic}} \\
    Network & MLP 256-128 \\
    Discount $\gamma$ / GAE $\lambda$ & 0.99 / 0.95 \\
    Epochs / minibatches & 5 / 4 \\
    Real-return windows & 2048 windows of 16 steps from the newest 200 steps, blocks with $\sigma\le0.2$ only \\
    \midrule
    \multicolumn{2}{@{}l}{\textbf{Exploration}} \\
    Action-noise blocks & $\sigma\in\{0,0.2,0.4,0.6\}$, AR(1) with $\rho=0.8$ \\
    Finger noise scale & 2.2 \\
    Observation noise & uniform, as in SuperMimic \\
    \midrule
    \multicolumn{2}{@{}l}{\textbf{Optimization}} \\
    Optimizer & Adam; policy $\beta=(0.9,0.95)$ \\
    Learning rate & $2\times10^{-4}$, cosine decay to 3\% over 250k iterations \\
    Updates per vectorized simulator step & 0.6 \\
    Iterations & 40k \\
    Precision / hardware / time & bf16 / one RTX 4090 / $\approx$\qty{15}{h} \\
    \bottomrule
  \end{tabular}
\end{table*}

\section{Experiments}
\label{sec:supp_grp_experiments}

This section first defines the evaluation of \cref{sec:exp} (\cref{sec:supp_grp_eval}), then gives the baselines
(\cref{sec:supp_grp_exp}), the training-data experiments (\cref{sec:supp_scaling}) and the ablations
(\cref{sec:supp_grp_abl}) in full, and finally repeats the main comparisons at the XHAND's real torque limits
(\cref{sec:supp_t1}).

\subsection{Evaluation protocol}
\label{sec:supp_grp_eval}

Zero-shot success is meaningful only if the zero-shot clips are unseen in training, so we first define the zero-shot
sets of \cref{sec:setup} and check them for leakage (\cref{sec:supp_zs}), then give the metrics and how we compute
intervals (\cref{sec:supp_eval}).

\subsubsection{Zero-shot sets}
\label{sec:supp_zs}

\begin{table}[!t]
  \centering
  \footnotesize
  \setlength{\tabcolsep}{3pt}
  \caption{\textbf{Zero-shot sets.} Clips: clips that pass the leakage checks below; Objects: objects of these
  clips; for EgoHOR, reconstructions, one per clip (\cref{sec:supp_count}).
  $^{a}$Identified in the dataset only by IDs.
  $^{b}$Eraser, toy, hammer, screwdriver and glue gun; only clips in which the object moves by at least
  \qty{2}{cm}, since a hand steadying a resting target succeeds trivially.
  $^{c}$Lid, remote, basket, can and shaker, and for GenHE-ZS also phone and cosmetics, which occur almost only in
  Gen-HumanEgo. The held-out lists of all sets are released with the code.}
  \label{tab:supp_zs}
  \fitcol{%
  \begin{tabular}{@{}llcrr@{}}
    \toprule
    Set & Held out from training & Hands & Clips & Objects \\
    \midrule
    HOT3D-ZS      & 6 objects$^{a}$                    & 2 & 192 & 6 \\
    OakInk2-ZS    & 18 categories$^{a}$                & 2 & 277 & 21 \\
    OakInk2-ZS-1H & the 18 of OakInk2-ZS               & 1 & 76  & 14 \\
    TACO-ZS       & 5 tool categories$^{b}$            & 2 & 178 & 19 \\
    DexYCB-ZS     & the whole dataset                  & 1 & 150 & 15 \\
    EgoStd-ZS     & 5 categories$^{c}$                 & 2 & 115 & 115 \\
    GenHE-ZS      & 7 categories$^{c}$                 & 2 & 235 & 235 \\
    \bottomrule
  \end{tabular}}
\end{table}

\Cref{tab:supp_zs} lists the zero-shot sets. Each holds out whole object categories, whole objects (HOT3D)
or a whole dataset (DexYCB): every clip of a held-out category or object is removed from training.

\paragraph{Categories.} For EgoHOR, an object's category is read from its name: the scene
identifier and words for side, color and size are dropped, synonyms are merged (\eg wastebasket $\to$ basket,
controller $\to$ remote, tin $\to$ can), and the object belongs to a held-out category if any remaining word
names it; objects with vague names (\eg ``object'') are used for training only. Phones and cosmetics occur
almost exclusively in Gen-HumanEgo; their clips from both sources are removed from training. For TACO,
an object's category follows from the tool and target names of its sequences; for OakInk2 we use its
object category identifiers and choose 18 categories greedily until the held-out clips reach the target size.

\paragraph{Leakage checks.} Three checks keep test clips apart from the training data. First, \emph{sibling
clips}, which share the human hand motion with a test clip (the two objects of one
reconstructed scene, the two halves of a split OakInk2 or TACO stage), are removed from training.

Second, \emph{near-duplicates}: every test object is compared with every training object after centering and
aligning both by their principal axes (four sign flips, metric scale kept; objects whose extents differ by
more than half are not compared); a symmetric mean surface distance below \qty{0.2}{cm} marks a
near-duplicate, and its test clips are not used. This detects copies of the same scanned mesh, such as DexYCB's scissors and DexCanvas's
scissors.

\begin{figure}[!t]
  \centering
  \includegraphics[width=\linewidth]{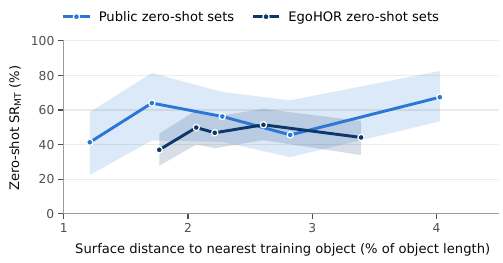}
  \caption{\textbf{Zero-shot success against the distance to the nearest training object}, as a percentage of the
  test object's length (its longest side; \cref{sec:supp_zs}). Each point is one fifth of the clips of the public or of
  the EgoHOR zero-shot sets, pooled over the sets and ordered by this distance (the near--far comparison in
  \cref{sec:supp_zs} is within each set). Bands: 95\% bootstrap intervals
  that resample objects. Final 4$\times$ policy, three seeds.}
  \label{fig:supp_nn}
\end{figure}

We do not use a larger threshold for near-duplicates: everyday objects share coarse shapes, so nearly every test object
has a training object of similar shape, whatever its category (after alignment, at a median of about 2\% of
the test object's longest side). Zero-shot success shows no sign of relying on such neighbors (\cref{fig:supp_nn}): within each set, the half of the clips whose objects are closest to a training object
succeeds less often than the other half, not more (by 9.1 points; 95\% interval $[0.8, 18.9]$ with objects resampled,
\cref{sec:supp_eval}). This does not rule out every effect of similar shapes or similar tasks, but clips with a closer
training object are not more likely to succeed.

Third, \emph{same category under a vague name}: an object of a held-out category can enter training under a vague
name, such as a cup lid named ``initial'' in a task about stacking cup lids. We do not use test clips whose
nearest training object comes from a clip whose task description names the test clip's category.

\subsubsection{Metrics and protocol}
\label{sec:supp_eval}

\paragraph{Metrics.} FDR, like several other trackers, trains with \emph{assistance}: a PD force and torque on the
object that pull it toward its reference pose, lowered by a curriculum (\cref{sec:supp_fdr}). Evaluation rollouts
apply no assistance, for every method. For SR$_{\text{MT}}$, the object position and rotation errors are each averaged
over the clip after a grace period of 15 steps (\qty{0.25}{s}). The hand-keypoint error compares the
proximal and distal finger links with the corresponding MANO joints. The normalized return is the episode return divided by the
largest attainable one. Compute is the wall-clock training time on one GPU, including compilation;
evaluation runs on other GPUs and is not counted.

\paragraph{Clips and repetitions.} The 563 public training clips are a fixed sample of the public training data,
drawn across all datasets. The 300 EgoHOR ones, 150 per source with at most one per scene or
recording, are trained on by every run whose data include their source. GPU nondeterminism changes the outcome of some clips, so every final checkpoint is
evaluated three times; learning curves evaluate intermediate checkpoints once. The tables give the range of the zero-shot mean
across seeds; that of the training-clip success is at most 3.7 points wide, except for two rows with an unstable seed
(decayed assistance at 4$\times$, 13.8; the Sharpa without the exploration design, 9.1).

\paragraph{Intervals.} Clips of one object are not independent: HOT3D-ZS, for example, holds 192 clips of only 6
objects (\cref{tab:supp_zs}), so resampling clips would overstate how much data a set holds. Our 95\% bootstrap
intervals therefore resample objects, each with all its clips. Each zero-shot set is resampled separately, and a
difference between two rows uses the same draw for both, so that the two are compared on the same objects; we take
2000 draws. Before resampling, each clip's success is averaged over seeds and evaluations. \Cref{tab:supp_ci} gives them for the zero-shot comparisons of \cref{tab:baselines,tab:ablation,tab:sets_t1}:
our policy outperforms reference replay and every learned baseline, is level with SPIDER at 4$\times$ and behind it at
1$\times$, and every ablation row except the one replacing the world model's object and scene inputs lowers zero-shot
success. Resampling clips instead gives narrower intervals and the same conclusions.

\begin{table}[!t]
  \centering
  \footnotesize
  \setlength{\tabcolsep}{3.5pt}
  \caption{\textbf{Lead of Ours in zero-shot success, with 95\% bootstrap intervals}: SR$_{\text{MT}}$ of Ours (mean of the
  seven zero-shot sets; 51.0\% at 4$\times$, 31.0\% at 1$\times$) minus that of the baselines and the ablation rows of
  \cref{tab:ablation} (\cref{tab:abl_sets,tab:sets_t1}), in points (objects resampled with all their clips, leads computed before rounding).
  Gray: interval contains 0.}
  \label{tab:supp_ci}
  \fitcol{%
  \begin{tabular}{@{}l r@{\;}l @{\hspace{12pt}} r@{\;}l@{}}
    \toprule
    & \multicolumn{2}{c}{4$\times$} & \multicolumn{2}{c}{1$\times$} \\
    \midrule
    Reference replay                        & 39.5 & $[35.0, 44.2]$ & 19.9 & $[16.8, 23.6]$ \\
    SPIDER                                  & \textcolor{gray}{1.6} & \textcolor{gray}{$[-2.1, 5.5]$} & $-$11.4 & $[-15.2, -6.9]$ \\
    SuperMimic                              & 37.2 & $[33.0, 41.4]$ & 19.2 & $[15.9, 22.7]$ \\
    FDR                                     & 33.0 & $[29.2, 36.8]$ & 15.9 & $[13.5, 18.6]$ \\
    \midrule
    \multicolumn{5}{@{}l}{\textit{Ours without}} \\
    \quad contact geometry $\phi$ (WM)      & 14.1 & $[11.8, 16.5]$ & 8.5 & $[6.5, 10.4]$ \\
    \quad object, scene inputs (WM, $\pi$)  & 17.8 & $[14.9, 20.7]$ & 8.8 & $[7.0, 10.6]$ \\
    \quad object, scene inputs (WM)         & \textcolor{gray}{$-$0.9} & \textcolor{gray}{$[-2.1, 0.4]$} & \textcolor{gray}{$-$0.7} & \textcolor{gray}{$[-1.9, 0.6]$} \\
    \quad object inputs ($\pi$)             & 16.3 & $[13.7, 19.1]$ & 9.3 & $[7.3, 11.2]$ \\
    \quad two-hand structure                & 7.8 & $[5.9, 9.6]$ & 4.8 & $[3.3, 6.3]$ \\
    \quad reward, training design           & 6.0 & $[4.1, 8.0]$ & 4.7 & $[3.0, 6.0]$ \\
    \quad exploration design                & 12.1 & $[10.4, 13.9]$ & 4.7 & $[3.1, 6.4]$ \\
    \bottomrule
  \end{tabular}}
\end{table}

\subsection{Comparison with baselines}
\label{sec:supp_grp_exp}

We detail three of the baselines of \cref{tab:baselines}, one for each kind of approach: a generalist tracker, FDR,
which we reimplement in our simulator (\cref{sec:supp_fdr}); per-clip optimization, SPIDER (\cref{sec:supp_spider});
and the world-model loop we build on, SuperMimic (\cref{sec:supp_supermimic}). \Cref{tab:base_sets} gives their
zero-shot success per set, next to that of the ablations, and \cref{sec:supp_t1} repeats the comparison at
1$\times$ torque. The fourth baseline, reference replay, plays back the retargeted joint targets without learned
correction (\cref{sec:exp_baselines}).

\begin{table*}[!t]
  \centering
  \footnotesize
  \setlength{\tabcolsep}{4.5pt}
  \caption{\textbf{Baselines and ablations at 4$\times$ torque}: SR$_{\text{MT}}$ (\%) on the training clips and per
  zero-shot set. Bold: best of Ours and the baselines in Train and Mean; red / gray: drop / gain relative to Ours;
  Seeds: range of the zero-shot mean over three seeds ($^{\ast}$two; --: one run); GPU-h: training (SPIDER: all zero-shot
  clips; ablation rows 13.1 to 18.2). FDR, assistance released: retrained with no assistance left after 300M steps
  (\cref{sec:supp_fdr}). Decayed assistance: one seed diverged (\cref{sec:supp_ablation}).}
  \label{tab:abl_sets}\label{tab:base_sets}\label{tab:spider_sets}
  \fitcol{%
  \begin{tabular}{@{}lr@{\hspace{1pt}}lrrrrrrrr@{\hspace{1pt}}lrr@{}}
    \toprule
    & & & \multicolumn{7}{c}{Zero-shot set (-ZS)} & & & & \\
    \cmidrule(lr){4-10}
    & \multicolumn{2}{c}{Train} & HOT3D & DexYCB & OakInk2 & OakInk2-1H & TACO & EgoStd & GenHE & \multicolumn{2}{c}{Mean} & Seeds & GPU-h \\
    \midrule
    Ours                                 & \textbf{69.4} & & 40.4 & 35.2 & 62.2 & 44.6 & 80.2 & 51.3 & 43.2 & \textbf{51.0} & & 49.6--53.5 & 15.3 \\
    \midrule
    \multicolumn{14}{@{}l}{\textit{Baselines} (\cref{tab:baselines} and \cref{sec:supp_fdr})} \\
    Reference replay                     & 19.7 & & 0.0  & 0.7  & 27.0 & 1.3  & 40.1 & 2.9  & 8.8  & 11.5 & & -- & -- \\
    SPIDER~\citesupp{pan2025spider}          & -- & & 52.1 & 20.7 & 72.9 & 59.2 & 72.5 & 31.3 & 37.0 & 49.4 & & -- & 111 \\
    SuperMimic~\citesupp{liao2026supermimic} & 28.2 & & 0.3  & 1.8  & 29.5 & 14.7 & 32.6 & 9.9  & 8.2  & 13.8 & & 13.4--14.3$^{\ast}$ & 13.9 \\
    FDR~\citesupp{lee2026fdr}                & 31.1 & & 0.0  & 10.1 & 29.7 & 9.4  & 60.0 & 6.9  & 9.9  & 18.0 & & 16.3--18.9 & 33.9 \\
    FDR, assistance released                 & 33.8 & & 0.2  & 7.9  & 30.5 & 8.8  & 60.3 & 6.6  & 9.3  & 17.6 & & 16.4--18.8 & 34.4 \\
    \midrule
    \multicolumn{14}{@{}l}{\textit{Ablations} (\cref{tab:ablation} and \cref{sec:supp_ablation})} \\
    Ours w/o contact geometry $\phi$ (world model) & 55.5 & \sdn{13.9} & 16.4 & 25.3 & 48.6 & 33.3 & 76.4 & 32.0 & 26.6 & 37.0 & \sdn{14.0} & 36.7--37.2$^{\ast}$ & \\
    Ours w/o object, scene inputs (world model, policy)        & 60.8 & \sdn{8.6} & 14.1 & 23.0 & 47.5 & 20.8 & 74.3 & 30.9 & 22.0 & 33.2 & \sdn{17.8} & 33.0--33.5$^{\ast}$ & \\
    Ours w/o object, scene inputs (world model) & 68.8 & \sdn{0.6} & 42.4 & 40.7 & 60.6 & 47.7 & 79.3 & 49.8 & 43.0 & 51.9 & \sgn{0.9} & 51.6--52.1 & \\
    Ours w/o object inputs (policy) & 61.7 & \sdn{7.7} & 14.1 & 26.6 & 46.9 & 28.1 & 71.6 & 33.2 & 22.3 & 34.7 & \sdn{16.3} & 32.9--36.5$^{\ast}$ & \\
    Ours w/o two-hand structure          & 63.3 & \sdn{6.1} & 25.5 & 24.9 & 56.7 & 42.8 & 78.3 & 41.2 & 33.3 & 43.2 & \sdn{7.8} & 42.9--43.5$^{\ast}$ & \\
    Ours w/o PointNet (policy)           & 66.0 & \sdn{3.4} & 20.7 & 31.2 & 53.1 & 28.7 & 75.3 & 40.5 & 29.0 & 39.8 & \sdn{11.2} & 39.3--40.5 & \\
    Ours w/o mirrored shared module      & 63.4 & \sdn{6.0} & 23.5 & 26.4 & 54.0 & 43.6 & 78.2 & 36.3 & 28.0 & 41.4 & \sdn{9.6} & 41.1--41.8 & \\
    Ours w/o reward, training design     & 64.3 & \sdn{5.1} & 33.8 & 35.4 & 56.4 & 37.3 & 73.6 & 45.5 & 33.1 & 45.0 & \sdn{6.0} & 44.9--45.2$^{\ast}$ & \\
    Ours w/o smoothness terms            & 71.9 & \sgn{2.5} & 47.1 & 49.6 & 65.4 & 50.1 & 80.6 & 55.4 & 49.3 & 56.8 & \sgn{5.8} & 55.8--58.4 & \\
    Ours w/o exploration design          & 63.3 & \sdn{6.1} & 26.2 & 15.0 & 54.4 & 32.7 & 75.1 & 38.4 & 30.5 & 38.9 & \sdn{12.1} & 38.5--39.3$^{\ast}$ & \\
    \quad only SuperMimic's noise          & 64.1 & \sdn{5.3} & 27.7 & 21.1 & 56.9 & 42.4 & 74.6 & 37.8 & 30.4 & 41.6 & \sdn{9.4} & 41.2--42.3 & \\
    \quad only w/o finger low-pass         & 67.9 & \sdn{1.5} & 36.5 & 27.7 & 61.9 & 40.8 & 78.7 & 44.8 & 41.3 & 47.4 & \sdn{3.6} & 46.6--48.2 & \\
    \quad only w/o return filter           & 67.0 & \sdn{2.4} & 35.4 & 36.7 & 61.0 & 42.3 & 81.7 & 46.0 & 38.9 & 48.8 & \sdn{2.2} & 47.7--49.6 & \\
    \quad one level $\sigma=0.3$, w/o return filter & 67.2 & \sdn{2.2} & 35.2 & 39.8 & 61.1 & 46.1 & 77.1 & 52.4 & 39.4 & 50.1 & \sdn{0.9} & 49.0--51.0 & \\
    Ours + object assistance, adaptive   & 20.5 & \sdn{48.9} & 0.2  & 5.3  & 30.5 & 8.1  & 52.8 & 8.6  & 8.9  & 16.3 & \sdn{34.7} & 14.0--18.6$^{\ast}$ & \\
    Ours + object assistance, decayed    & 65.5 & \sdn{3.9} & 33.7 & 39.7 & 61.6 & 40.1 & 78.8 & 45.4 & 38.9 & 48.3 & \sdn{2.7} & 42.3--51.5 & \\
    \bottomrule
  \end{tabular}}
\end{table*}

\subsubsection{Generalist tracker (FDR)}
\label{sec:supp_baselines}\label{sec:supp_fdr}

We compare with FDR because, in its own study on the XHAND, it outperforms per-clip RL
(DexMachina~\citesupp{mandi2025dexmachina}, CHORD~\citesupp{zhu2026chord}) and sampling-based optimization (Do as I
Do~\citesupp{paliwal2026doasido}) by a wide margin and at lower compute; the same study finds that PPO learns little
on this task and attributes FDR's results to its off-policy agent, FlashSAC.
We reimplement this agent from the released code, together with its observations, rewards, terminations,
assistance curriculum and reset sampling, so it differs from ours in the method, not in the task.
We validate it in FDR's original bimanual setting (below). On our training data, we train it for 600M steps (about
34 GPU-hours) with three seeds.
FDR's published numbers come from its own simulator and physics settings, and count a training clip as converted
once any rollout during training tracks it. Here all methods run in the same simulator and physics, and we
evaluate the final policy on fresh rollouts: zero-shot clips are never trained on, so they can only be scored
this way, and we score the training clips the same way.

\paragraph{Differences from the original implementation.}
The agent (FlashSAC), its replay buffer and its trajectory encoder are FDR's released code, used unchanged with its
hyperparameters. The rest is rewritten for our simulator:
\begin{itemize}
  \item \emph{Physics}: ours, which is the purpose of the reimplementation.
  \item \emph{Contact observations}: the same per-link contact force, point and torque as FDR's PhysX backend, with
    each link's force taken from our hand--object contact sensor.
  \item \emph{Single-hand clips}, which FDR does not support: the samples of the idle hand get no reward or
    termination penalty and are left out of the losses and the normalization statistics; with bimanual clips only,
    this reduces to FDR's training.
  \item \emph{Reset sampling}: FDR draws a clip uniformly and a start frame uniformly from its first 99\%. Since each of
    our environments holds one object, we draw an object in proportion to its number of clips and then one of its
    clips, which gives the same distribution.
\end{itemize}

\paragraph{Validation.} We train the reimplementation in FDR's original bimanual setting (1540 clips with support
surfaces, 300M environment steps). After 300M steps it reaches 28.6\% success, against 28.6 and 31.2\% for the two
seeds of the original implementation; in mid-training it lags slightly, because its curriculum lowers the assistance
a little later.

\paragraph{Assistance curriculum on our data.} FDR lowers the gains of its assistance by 10\% only when, among
other conditions, the completion rate averaged over 30 epochs reaches 0.65. On our data it stays at 0.61 to 0.64, so
the assistance is never fully released: after 600M steps, the upper bound of the position gain is 7.1 at 4$\times$
and 10.8 to 12.0 at 1$\times$, 9\% and 14\% of its initial value of 80. In FDR's own setting it is 1.5 after 300M
steps (2\%).
To test whether the assistance left at the end explains FDR's results, we retrain FDR with the curriculum's
gain bounds additionally capped at their initial values times $1-t/\text{300M}$ for step $t$, so that no assistance
is left after 300M steps (600M steps; \cref{tab:abl_sets,tab:sets_t1}). Evaluated without assistance, the retrained
FDR reaches about the same zero-shot success as the original (17.6 \vs 18.0\% at 4$\times$, with overlapping seed
ranges, and 15.8 \vs 15.2\% at 1$\times$) and about 3 points more on the training clips. The assistance its curriculum leaves therefore
does not explain FDR's results.

\subsubsection{Per-clip optimization (SPIDER)}
\label{sec:supp_spider}

We use SPIDER's OakInk exploration parameters, 32 optimizer iterations per
control step and no early stopping. It runs on the same MuJoCo model that our simulator steps, exported per clip,
with the same physics as our policy: finger torque limits, the gravity-compensated and limited wrist, and the
elliptic friction cone. This costs about 326 GPU-seconds per clip and 111 GPU-hours for the zero-shot clips, about seven times our training. Its results per set are in
\cref{tab:base_sets}.

\subsubsection{World-model loop (SuperMimic)}
\label{sec:supp_supermimic}

Applied to our task, SuperMimic~\citesupp{liao2026supermimic} keeps its own hyperparameters and uses no
contact geometry, features of the support surfaces instead of scene distances, a learned embedding per
object, no point cloud for the policy, small uncorrelated Gaussian action noise, one policy on the concatenated inputs of both hands, no
smoothness terms, and none of our reward and training design choices. It trains for 40k iterations in 13.9
GPU-hours (ours: 15.3). Its hyperparameters explain only a small part of the gap: the ablation row without our
reward and training design uses SuperMimic's horizon, world-model batches and update ratio and none of our reward and
training changes, but keeps our inputs, two-hand structure and exploration, and still reaches 45.0\% zero-shot,
against 13.8\% for SuperMimic (\cref{tab:abl_sets}).

\subsection{Effect of the amount and source of training data}
\label{sec:supp_scaling}

We detail the training-data experiments of \cref{sec:exp_scaling} and give them in full; their results at 1$\times$
torque are discussed in \cref{sec:supp_t1}.

\paragraph{Data scaling.}
The subsets of \cref{fig:scaling} are nested and drawn by object. Zero-shot success rises with the data on every set,
most on HOT3D-ZS, DexYCB-ZS and OakInk2-ZS-1H; only TACO-ZS is slightly higher at half of the data than with all of it.

The training-clip curve of \cref{fig:scaling}a also reflects generalization. The smaller subsets contain few of its
clips: of the 563 public ones, 97, 85 and 62\% are missing at 12.5, 25 and 50\% (they come from only 70 objects). On
the 349 public clips that none of these subsets contains, success still rises with the data (71.4, 75.7 and 77.9\%),
so more data also improves the conversion of clips a policy was not trained on.

\paragraph{Adding EgoHOR.} The runs of \cref{tab:ego} use nested subsets: EgoHOR by scene (EgoStandard) or recording
(Gen-HumanEgo), keeping sibling and repaired clips with their originals, and the public data by object, with the same
share of every dataset. When EgoHOR is present it receives a fixed half
of the bimanual environments. Each row trains for 40k iterations with three seeds. \Cref{tab:supp_ego} gives every
row at 4$\times$ and 1$\times$ torque, together with the smaller subsets of \cref{fig:scaling}a, with the two EgoHOR
sets separately and the success on the training clips of the full data: its 563 public clips, its 300 EgoHOR training
clips (\cref{sec:supp_eval}) and all 863 of them (as in \cref{fig:scaling}a).

The training clips show what adding EgoHOR costs. On the public ones, almost nothing: with all public data, success
is 80.3\% without EgoHOR and 79.5\% with it. On the 300 EgoHOR training clips, which every EgoHOR subset keeps,
success is higher with 12.5\% of EgoHOR than with all of it (55.5 \vs 50.3\%), presumably because its fixed share of
the environments then trains fewer clips; with more scenes and objects to learn from, more EgoHOR instead raises
success on the EgoHOR zero-shot sets (39.7, 42.6 and 47.2\% for none, 12.5\% and all of it; \cref{tab:supp_ego}).

\begin{table*}[!t]
  \centering
  \footnotesize
  \setlength{\tabcolsep}{3.5pt}
  \caption{\textbf{Effect of the amount and source of training data} at 4$\times$ and 1$\times$ torque: all rows of
  \cref{tab:ego} and the smaller subsets of all data of \cref{fig:scaling}a (last three rows). SR$_{\text{MT}}$ (\%); Share:
  of the public data (Pub.) and of EgoHOR (Ego.). Training clips: those of the full data, 563 public, 300 EgoHOR and all
  863; $^\dagger$none of them is in this row's training data; with less than all of the public data, most public ones are
  not either (\cref{sec:supp_scaling}). Zero-shot Pub. / All: mean over the five public / all seven sets. Bold: best zero-shot mean per torque.}
  \label{tab:supp_ego}
  \fitcol{%
  \begin{tabular}{@{}rr@{\hspace{16pt}}ccc ccccc@{\hspace{24pt}}ccc ccccc@{}}
    \toprule
    & & \multicolumn{8}{c}{4$\times$ torque} & \multicolumn{8}{c}{1$\times$ torque} \\
    \cmidrule(lr){3-10}\cmidrule(l){11-18}
    \multicolumn{2}{c}{Share (\%)} & \multicolumn{3}{c}{Training clips} & \multicolumn{5}{c}{Zero-shot} &
      \multicolumn{3}{c}{Training clips} & \multicolumn{5}{c}{Zero-shot} \\
    \cmidrule(r){1-2}\cmidrule(lr){3-5}\cmidrule(lr){6-10}\cmidrule(lr){11-13}\cmidrule(l){14-18}
    & & & & & & \multicolumn{3}{c}{EgoHOR} & & & & & & \multicolumn{3}{c}{EgoHOR} & \\
    \cmidrule(lr){7-9}\cmidrule(lr){15-17}
    Pub. & Ego. & Pub. & Ego. & All & Pub. & EgoStd & GenHE & Mean & All & Pub. & Ego. & All & Pub. & EgoStd & GenHE & Mean & All \\
    \midrule
     100 &  100 & 79.5 & 50.3 & 69.4 & \textbf{52.5} & 51.3 & 43.2 & \textbf{47.2} & \textbf{51.0} & 65.3 & 36.4 & 55.2 & \textbf{32.0} & 31.2 & 26.0 & \textbf{28.6} & \textbf{31.0} \\
     100 & 12.5 & 78.9 & 55.5 & 70.8 & 51.4 & 47.2 & 37.9 & 42.6 & 48.9 & 65.4 & 39.4 & 56.3 & 31.1 & 27.1 & 23.4 & 25.3 & 29.5 \\
     100 &    0 & 80.3 & 39.3$^\dagger$ & 66.1 & 51.6 & 43.5 & 36.0 & 39.7 & 48.2 & 66.5 & 25.9$^\dagger$ & 52.4 & 31.3 & 22.8 & 19.1 & 21.0 & 28.4 \\
      50 &  100 & 76.5 & 50.7 & 67.5 & 50.5 & 47.6 & 40.7 & 44.2 & 48.7 & 63.2 & 34.4 & 53.2 & 30.2 & 31.8 & 24.5 & 28.1 & 29.6 \\
      50 &    0 & 78.6 & 37.2$^\dagger$ & 64.2 & 47.2 & 41.7 & 33.3 & 37.5 & 44.5 & 64.6 & 24.0$^\dagger$ & 50.5 & 29.4 & 18.6 & 17.4 & 18.0 & 26.1 \\
      25 &  100 & 73.7 & 49.8 & 65.4 & 48.5 & 46.2 & 40.9 & 43.5 & 47.1 & 58.8 & 33.6 & 50.1 & 27.4 & 28.2 & 22.4 & 25.3 & 26.8 \\
      25 &    0 & 73.8 & 33.2$^\dagger$ & 59.7 & 43.1 & 36.0 & 29.6 & 32.8 & 40.2 & 60.4 & 21.2$^\dagger$ & 46.7 & 27.0 & 18.8 & 14.7 & 16.7 & 24.1 \\
    \addlinespace
      50 &   50 & 77.5 & 51.5 & 68.5 & 51.4 & 49.0 & 39.9 & 44.4 & 49.4 & 62.4 & 35.7 & 53.1 & 30.2 & 30.4 & 22.6 & 26.5 & 29.1 \\
      25 &   25 & 74.1 & 52.3 & 66.5 & 47.4 & 46.6 & 40.6 & 43.6 & 46.3 & 58.9 & 36.2 & 51.0 & 26.9 & 27.0 & 21.3 & 24.1 & 26.1 \\
    12.5 & 12.5 & 68.6 & 53.3 & 63.3 & 43.3 & 44.3 & 35.7 & 40.0 & 42.4 & 54.1 & 37.6 & 48.3 & 24.3 & 25.3 & 19.5 & 22.4 & 23.8 \\
    \bottomrule
  \end{tabular}}
\end{table*}

\subsection{Ablations and analysis}
\label{sec:supp_grp_abl}

Each row of \cref{tab:ablation} replaces one group of components, or part of a group. We state what each row changes, add rows that
isolate single components, the smoothness terms and object assistance (\cref{sec:supp_ablation}), and relate the world
model's prediction error to the policy's success (\cref{sec:supp_wm_analysis}).

\subsubsection{Ablation rows and further ablations}
\label{sec:supp_ablation}

\paragraph{Ablation rows.} \Cref{tab:abl_rows} lists what each row of \cref{tab:ablation} changes. Every row keeps
the task, its physics and the data of Ours, and changes only the listed components. Some rows appear only here
(\cref{tab:abl_sets}): those that isolate the PointNet, the mirrored module and the smoothness terms, the four that
split the exploration group, and the two with object assistance. The columns of \cref{tab:abl_rows}, with what a row
uses instead (\,$\to$):
\begin{itemize}\setlength{\itemsep}{0pt}\setlength{\parskip}{0pt}
  \item $\phi$: contact geometry (removed);
  \item $z_o$: object code (\,$\to$ a learned per-object embedding);
  \item $\psi$: scene distances (\,$\to$ parameters of the support surfaces);
  \item Frames: gravity in the wrist frame, no horizontal wrist position (\,$\to$ world frame);
  \item Encoding: the policy's hand-relative PointNet, its contact geometry $\phi^h$ and the local probes $\chi^h$
    (\,$\to$ FDR's flat point cloud with the palm and fingertip distances only), changed together;
  \item Two-hand: one mirrored module for both hands with the hands' activity flags; a critic without the second
    hand's flat point cloud and with the idle hand's features zeroed; half of the environments on bimanual and half
    on single-hand clips, objects weighted by the square root of their frames, with fixed source shares
    (\,$\to$ one network on both hands' inputs, the critic on both hands' full features, environments over all
    clips by frames), changed together;
  \item Reward, training: squared kernels, rotation term, weight EMA, real-return critic targets, horizon
    24, 16 world-model batches, update ratio 0.6 (\,$\to$ FDR's kernels; no rotation term, weight EMA or real-return
    targets; SuperMimic's 32, 10 and 0.4);
  \item Blocks: four AR(1) noise levels (\,$\to$ SuperMimic's uncorrelated Gaussian $\sigma=0.1$, or one level of 0.3);
  \item Filter: return filter, the critic's real-return windows only from the blocks with $\sigma\le0.2$, whose returns
    reflect the policy rather than the noise; Low-pass: finger low-pass with scaled finger noise;
  \item No assist.: no assistance on the object (\,$\to$ FDR's assistance, with its gains as a world-model input,
    lowered by its adaptive curriculum or decayed linearly to zero over the first 20k iterations).
\end{itemize}

\begin{table*}[!t]
  \centering
  \footnotesize
  \setlength{\tabcolsep}{4pt}
  \caption{\textbf{Ablation rows}: \checkmark\ as in Ours, $\times$ replaced or removed (columns in
  \cref{sec:supp_ablation}). $^{\dagger}$In \cref{tab:abl_sets} only. $^{a}$Only the PointNet, replaced by FDR's
  flat point cloud; contact geometry $\phi^h$ and local probes $\chi^h$ kept. $^{b}$Only the mirrored shared module, replaced by one
  network on both hands' inputs; critic inputs and environment allocation as in Ours. $^{c}$Only the smoothness
  terms $r_{\text{vel}}$, $r_{\Delta\text{w}}$, $r_{\text{jerk}}$. $^{d}$SuperMimic's uncorrelated Gaussian $\sigma=0.1$, which leaves
  the return filter nothing to select.}
  \label{tab:abl_rows}
  \begin{tabular}{@{}l cccc cc c ccc c@{}}
    \toprule
    & \multicolumn{4}{c}{World model (\cref{sec:wm})} & \multicolumn{2}{c}{Policy (\cref{sec:policy})} &
      Reward, & \multicolumn{3}{c}{Exploration (\cref{sec:training})} & \\
    \cmidrule(lr){2-5} \cmidrule(lr){6-7} \cmidrule(lr){9-11}
    Row & $\phi$ & $z_o$ & $\psi$ & Frames & Encoding & Two-hand & training & Blocks & Filter & Low-pass & No assist. \\
    \midrule
    Ours                                        & \checkmark & \checkmark & \checkmark & \checkmark & \checkmark & \checkmark & \checkmark & \checkmark & \checkmark & \checkmark & \checkmark \\
    \midrule
    w/o contact geometry $\phi$ (world model)   & $\times$ & \checkmark & \checkmark & \checkmark & \checkmark & \checkmark & \checkmark & \checkmark & \checkmark & \checkmark & \checkmark \\
    w/o object, scene inputs (world model, policy)                    & \checkmark & $\times$ & $\times$ & $\times$ & $\times$ & \checkmark & \checkmark & \checkmark & \checkmark & \checkmark & \checkmark \\
    w/o object, scene inputs (world model)                      & \checkmark & $\times$ & $\times$ & $\times$ & \checkmark & \checkmark & \checkmark & \checkmark & \checkmark & \checkmark & \checkmark \\
    w/o object inputs (policy)                           & \checkmark & \checkmark & \checkmark & \checkmark & $\times$ & \checkmark & \checkmark & \checkmark & \checkmark & \checkmark & \checkmark \\
    w/o two-hand structure                      & \checkmark & \checkmark & \checkmark & \checkmark & \checkmark & $\times$ & \checkmark & \checkmark & \checkmark & \checkmark & \checkmark \\
    w/o reward, training design                 & \checkmark & \checkmark & \checkmark & \checkmark & \checkmark & \checkmark & $\times$ & \checkmark & \checkmark & \checkmark & \checkmark \\
    w/o exploration design                      & \checkmark & \checkmark & \checkmark & \checkmark & \checkmark & \checkmark & \checkmark & $\times$ & $\times$ & $\times$ & \checkmark \\
    \midrule
    w/o PointNet (policy)$^{\dagger}$           & \checkmark & \checkmark & \checkmark & \checkmark & $\times^{a}$ & \checkmark & \checkmark & \checkmark & \checkmark & \checkmark & \checkmark \\
    w/o mirrored shared module$^{\dagger}$      & \checkmark & \checkmark & \checkmark & \checkmark & \checkmark & $\times^{b}$ & \checkmark & \checkmark & \checkmark & \checkmark & \checkmark \\
    w/o smoothness terms$^{\dagger}$            & \checkmark & \checkmark & \checkmark & \checkmark & \checkmark & \checkmark & $\times^{c}$ & \checkmark & \checkmark & \checkmark & \checkmark \\
    only SuperMimic's noise$^{\dagger}$         & \checkmark & \checkmark & \checkmark & \checkmark & \checkmark & \checkmark & \checkmark & $\times^{d}$ & $\times$ & \checkmark & \checkmark \\
    only w/o finger low-pass$^{\dagger}$        & \checkmark & \checkmark & \checkmark & \checkmark & \checkmark & \checkmark & \checkmark & \checkmark & \checkmark & $\times$ & \checkmark \\
    only w/o return filter$^{\dagger}$          & \checkmark & \checkmark & \checkmark & \checkmark & \checkmark & \checkmark & \checkmark & \checkmark & $\times$ & \checkmark & \checkmark \\
    one level $\sigma=0.3$, w/o return filter$^{\dagger}$ & \checkmark & \checkmark & \checkmark & \checkmark & \checkmark & \checkmark & \checkmark & $\times$ & $\times$ & \checkmark & \checkmark \\
    + object assistance, adaptive$^{\dagger}$   & \checkmark & \checkmark & \checkmark & \checkmark & \checkmark & \checkmark & \checkmark & \checkmark & \checkmark & \checkmark & $\times$ \\
    + object assistance, decayed$^{\dagger}$    & \checkmark & \checkmark & \checkmark & \checkmark & \checkmark & \checkmark & \checkmark & \checkmark & \checkmark & \checkmark & $\times$ \\
    SuperMimic                                  & $\times$ & $\times$ & $\times$ & $\times$ & $\times$ & $\times$ & $\times$ & $\times$ & $\times$ & $\times$ & \checkmark \\
    \bottomrule
  \end{tabular}
\end{table*}

\paragraph{Isolating the PointNet and the mirrored module.}
The encoding and two-hand rows each change several components, so two more rows isolate one component each (three
seeds; \cref{tab:abl_rows,tab:abl_sets}). Replacing only the policy's PointNet by FDR's flat point cloud, with its
contact geometry $\phi^h$ and local probes $\chi^h$ kept, lowers zero-shot success by 11.2 points (to 39.8\%) but training-clip success by only 3.4 points; the zero-shot drop is about two thirds of that of the encoding row (16.3 points). Removing only the mirrored shared module, with the
critic's inputs and the environment allocation as in Ours, lowers zero-shot success by 9.6 points (to 41.4\%),
more than the whole two-hand row (7.8 points), and training-clip success by 6.0 points. The seed ranges of both rows (39.3--40.5 and
41.1--41.8\%) lie well below that of Ours (49.6--53.5\%).

\paragraph{Splitting the exploration design.}
The exploration group changes three things relative to SuperMimic: the noise (four blocks of temporally correlated
AR(1) noise with $\sigma$ up to 0.6, instead of uncorrelated Gaussian noise with $\sigma=0.1$), the return filter
(real-return critic targets only from the blocks with $\sigma\le0.2$), and the finger low-pass (with the finger noise
scaled up to make up for its attenuation). Removing all three costs 12.1 zero-shot points. Four rows change one part
each (\cref{tab:abl_rows,tab:abl_sets}):
\begin{itemize}\setlength{\itemsep}{0pt}\setlength{\parskip}{0pt}
  \item Noise: SuperMimic's noise, which also leaves the return filter nothing to select, costs 9.4 points, about
    three quarters of the group.
  \item Finger low-pass: removing it costs 3.6 points. All three seeds of these two rows lie below those of Ours.
  \item Return filter: removing it, so that the critic also learns from the returns of the high-noise blocks, costs
    2.2 points.
  \item Number of levels: without the return filter, a single AR(1) level of $\sigma=0.3$ (50.1\%) does no worse than
    the four levels (48.8\%).
\end{itemize}
What matters is thus replacing SuperMimic's small, uncorrelated noise by larger, temporally correlated noise; the
low-pass adds a few points, and the four levels with the return filter give no clear gain over a single level (0.9
points, within the seed variation).

\paragraph{Smoothness terms.}
The smoothness terms of \cref{eq:reward} trade some tracking success for smoother motion. We measure, on each run's
own successful clips, the content above \qty{5}{\hertz} of the finger targets and of the wrist rotation targets, and
the angular jitter of the palm (\cref{tab:quality}). FDR's finger targets carry about 5 times
as much of this content as those of our policy, and its wrist and palm about 2.5 times as much. Removing the
terms (three seeds) raises zero-shot success by 5.8 points (56.8\%, seeds 55.8--58.4\%) and success on the training
clips by 2.5 points, but the finger targets then carry about three times as much content above \qty{5}{\hertz}, the wrist
rotation targets and the palm about twice as much, the palm close to FDR (\cref{tab:quality}; the same ratios hold for
the mean over three seeds). On the clips that both runs succeed on (631 with FDR, about 1490 with each seed of the run
without the terms), the ratios are the same, and all three measures are lower for our policy on 99 to 100\% of these clips.
We keep the terms, since the trajectories are intended for execution on robots.

\begin{table}[!t]
  \centering
  \footnotesize
  \setlength{\tabcolsep}{4pt}
  \caption{\textbf{Smoothness} of one seed per method on its own successful clips: medians over clips of the content
  above \qty{5}{\hertz} of the finger targets (rad/s) and of the wrist rotation targets (\textdegree/s), and of the
  palm's angular jitter (\textdegree/s). Success: object criterion of SR$_{\text{MT}}$, mean of the seven zero-shot
  sets, one evaluation. 1$\times$ rows: \cref{sec:supp_t1}.}
  \label{tab:quality}
  \fitcol{%
  \begin{tabular}{@{}ll rrrr@{}}
    \toprule
    Torque & Method & Success & Fingers & Wrist & Palm \\
    \midrule
    4$\times$ & Ours & 55.8 & \textbf{4.1} & \textbf{20.4} & \textbf{9.8} \\
              & Ours w/o smoothness terms & 59.9 & 12.9 & 38.8 & 21.6 \\
              & FDR  & 17.1 & 19.3 & 50.1 & 24.8 \\
    \midrule
    1$\times$ & Ours & 34.1 & \textbf{2.4} & \textbf{19.0} & \textbf{9.3} \\
              & FDR  & 14.9 & 19.9 & 48.2 & 21.8 \\
    \bottomrule
  \end{tabular}}
\end{table}

\paragraph{Object assistance.} Several trackers apply assistance on the object early in
training~\citesupp{lee2026fdr,mandi2025dexmachina,zhu2026chord}. It brings no benefit in our framework, so we train
without it (\cref{tab:abl_sets}). With FDR's adaptive curriculum, our runs never meet the conditions under which it
lowers the assistance, so the policy learns to rely on forces it does not receive at evaluation (16.3\% zero-shot).
Decayed linearly to zero over the first 20k iterations, the assistance brings no gain either: two seeds end level with
Ours, and in the third the world model diverged early in training, so the mean is lower (48.3\%, against 51.0\%).

\subsubsection{Analysis of the ablations' world models}
\label{sec:supp_wm_analysis}

\paragraph{Setup.} To analyze the effect of the world model's inputs, every world model of \cref{tab:wm_err} predicts the same
simulator transitions: those of our final policy (first seed) on 600 training clips (300 public, 300 EgoHOR) and on the 1223
zero-shot clips, two rollouts per clip (one without and one with the training noise). From a recorded state, each
model predicts 24 steps (\qty{0.4}{s}) open loop with the recorded actions, in windows that end before the clip
terminates. Each row averages the world models of its seeds at 40k iterations (three for Ours and the row without the
object and scene inputs in the world model, two for the others). All world models are evaluated on transitions of the first seed of Ours, but each was trained on those of
its own policy, so a world model from another run has larger errors even with our inputs: those of the other two seeds
of Ours are up to 10\% above those of the first, and those of the row without the policy's object inputs, whose world
model has our inputs, 4 to 11\% above those of Ours. Differences of this size therefore do not reflect the world
model's inputs.

\begin{table}[!t]
  \centering
  \footnotesize
  \setlength{\tabcolsep}{2.5pt}
  \setlength{\fboxsep}{0.8pt}
  \caption{\textbf{World-model prediction error} of the object after 24 open-loop steps on the same transitions of our
  policy (\cref{sec:supp_wm_analysis}), mean over seeds. Ratio: zero-shot (ZS) / training clips. Next to each error: its
  increase over Ours; gray: up to that of a run with the same world model as Ours (11\%), red: above. WM: world model,
  $\pi$: policy. $^{a}$Same world model as Ours. $^{b}$Per-object identity embedding; unseen objects get the mean
  embedding. Bold: lowest error.}
  \label{tab:wm_err}
  \fitcol{%
  \begin{tabular}{@{}l r@{\hspace{1pt}}l r@{\hspace{1pt}}l r r@{\hspace{1pt}}l r@{\hspace{1pt}}l r@{}}
    \toprule
    & \multicolumn{5}{c}{Position (cm)} & \multicolumn{5}{c}{Rotation (\textdegree)} \\
    \cmidrule(lr){2-6} \cmidrule(l){7-11}
    World model of & \multicolumn{2}{c}{Train} & \multicolumn{2}{c}{ZS} & Ratio & \multicolumn{2}{c}{Train} & \multicolumn{2}{c}{ZS} & Ratio \\
    \midrule
    Ours & \textbf{1.28} &  & \textbf{2.03} &  & 1.59 & \textbf{12.8} &  & \textbf{15.9} &  & 1.24 \\
    \multicolumn{11}{@{}l}{\textit{Ours without}} \\
    \quad object inputs ($\pi$)$^{a}$ & 1.42 & \serrok{11} & 2.25 & \serrok{11} & 1.58 & 13.8 & \serrok{8} & 16.6 & \serrok{4} & 1.20 \\
    \quad object, scene (WM)$^{b}$ & 1.40 & \serrok{9} & 2.95 & \serrhi{45} & 2.11 & 13.6 & \serrok{6} & 16.2 & \serrok{2} & 1.19 \\
    \quad object, scene (WM, $\pi$)$^{b}$ & 1.47 & \serrhi{15} & 2.99 & \serrhi{47} & 2.03 & 14.1 & \serrok{10} & 17.0 & \serrok{7} & 1.21 \\
    \quad contact geometry $\phi$ (WM) & 2.18 & \serrhi{70} & 3.69 & \serrhi{82} & 1.69 & 16.8 & \serrhi{31} & 21.3 & \serrhi{34} & 1.26 \\
    \midrule
    SuperMimic & 3.50 & \serrhi{173} & 5.42 & \serrhi{167} & 1.55 & 21.8 & \serrhi{70} & 26.7 & \serrhi{68} & 1.23 \\
    \bottomrule
  \end{tabular}}
\end{table}

\paragraph{Contact geometry.} SuperMimic's world model, with none of our inputs, has by far the largest errors, and of
our inputs the contact geometry matters most: without it, the position error grows by 70 to 82\% and the rotation
error by 31 to 34\%, on training and zero-shot clips alike, so the contact geometry makes the world model more accurate
on seen and unseen objects.

\paragraph{Object and scene inputs.} With a learned per-object identity embedding in place of our object and scene inputs, the position error on the
training clips stays within this range ($+9$\%), but on the zero-shot clips it is 45\% larger: the embedding
identifies the training objects without describing their shape. With the policy's object inputs also replaced, the
position error is 15\% above ours on the training clips and 47\% on the zero-shot clips.

\paragraph{What matters for success.} Together with \cref{tab:abl_sets}, this shows what the world model needs. Its object and scene inputs only have to
make it accurate on the training objects: it is used only in training, on these objects, and not when the policy is
evaluated, so its larger errors on new objects leave success unchanged (51.9 \vs 51.0\% zero-shot, 68.8 \vs 69.4\% on
the training clips). Success depends instead on the policy's object inputs, without which zero-shot success drops by
16.3 points (17.8 with the world model's also replaced), and on the world model's contact geometry, without which its
errors grow on the training objects too and success drops by 14.0 points.

\subsection{Results at 1$\times$ torque}
\label{sec:supp_t1}

The main results use four times the XHAND's finger torque limits (\cref{sec:supp_sim}). At the real limits
(1$\times$), \cref{tab:sets_t1} repeats the comparison with the baselines and the ablation rows of
\cref{tab:ablation} (three seeds each). The training-data experiments and the smoothness at 1$\times$ share their
tables with those at 4$\times$ (\cref{tab:supp_ego,tab:quality}) and are discussed here; FDR retrained with its
assistance released is discussed with FDR (\cref{sec:supp_fdr}).

\begin{table*}[!t]
  \centering
  \footnotesize
  \setlength{\tabcolsep}{4pt}
  \caption{\textbf{Baselines and ablations at 1$\times$ torque}, as in \cref{tab:abl_sets}, for the rows of
  \cref{tab:ablation} and the two noise variants; the ablation rows train for 13.1 to 18.4 GPU-hours.}
  \label{tab:sets_t1}
  \fitcol{%
  \begin{tabular}{@{}lr@{\hspace{1pt}}lrrrrrrrr@{\hspace{1pt}}lrr@{}}
    \toprule
    & & & \multicolumn{7}{c}{Zero-shot set (-ZS)} & & & & \\
    \cmidrule(lr){4-10}
    & \multicolumn{2}{c}{Train} & HOT3D & DexYCB & OakInk2 & OakInk2-1H & TACO & EgoStd & GenHE & \multicolumn{2}{c}{Mean} & Seeds & GPU-h \\
    \midrule
    Ours                                 & \textbf{55.2} & & 13.5 & 10.1 & 48.4 & 25.3 & 62.9 & 31.2 & 26.0 & 31.0 & & 31.0--31.1 & 15.5 \\
    \midrule
    \multicolumn{14}{@{}l}{\textit{Baselines}} \\
    Reference replay                     & 19.7 & & 0.0  & 0.7  & 25.0 & 3.1  & 38.2 & 2.3  & 8.5  & 11.1 & & -- & -- \\
    SPIDER~\citesupp{pan2025spider}          & --   & & 35.9 & 14.0 & 69.0 & 61.8 & 65.2 & 19.1 & 31.9 & \textbf{42.4} & & -- & 110 \\
    SuperMimic~\citesupp{liao2026supermimic} & 23.9 & & 0.1  & 2.1  & 26.1 & 10.7 & 26.8 & 8.3  & 8.5  & 11.8 & & 11.3--12.2 & 14.1 \\
    FDR~\citesupp{lee2026fdr}                & 27.0 & & 0.1  & 4.4  & 31.3 & 6.3  & 48.1 & 6.2  & 9.7  & 15.2 & & 14.9--15.4 & 33.9 \\
    FDR, assistance released                 & 30.1 & & 0.0  & 3.6  & 30.6 & 8.3  & 51.2 & 6.7  & 10.4 & 15.8 & & 15.6--16.3 & 34.6 \\
    \midrule
    \multicolumn{14}{@{}l}{\textit{Ablations}} \\
    Ours w/o contact geometry $\phi$ (world model)       & 44.4 & \sdn{10.8} & 5.7 & 5.9 & 41.5 & 19.2 & 53.0 & 15.9 & 16.9 & 22.6 & \sdn{8.4} & 21.5--23.5 & \\
    Ours w/o object, scene inputs (world model, policy)  & 48.8 & \sdn{6.4} & 3.7 & 6.3 & 38.1 & 16.4 & 58.1 & 19.8 & 13.2 & 22.2 & \sdn{8.8} & 21.7--23.0 & \\
    Ours w/o object, scene inputs (world model)          & 55.6 & \sgn{0.4} & 13.6 & 12.0 & 47.9 & 27.2 & 65.2 & 30.8 & 25.2 & 31.7 & \sgn{0.7} & 30.0--33.3 & \\
    Ours w/o object inputs (policy)      & 47.9 & \sdn{7.3} & 4.7  & 5.2  & 38.9 & 13.7 & 58.8 & 17.6 & 13.5 & 21.8 & \sdn{9.2} & 20.1--23.1 & \\
    Ours w/o two-hand structure          & 50.6 & \sdn{4.6} & 6.5  & 7.7  & 44.4 & 21.8 & 60.7 & 25.1 & 17.5 & 26.3 & \sdn{4.7} & 24.7--27.6 & \\
    Ours w/o reward, training design     & 49.8 & \sdn{5.4} & 10.0 & 9.9  & 45.6 & 21.2 & 53.7 & 24.5 & 19.8 & 26.4 & \sdn{4.6} & 26.3--26.4 & \\
    Ours w/o exploration design          & 51.2 & \sdn{4.0} & 8.0  & 6.9  & 45.0 & 20.6 & 63.4 & 21.2 & 19.5 & 26.4 & \sdn{4.6} & 25.3--27.4 & \\
    \quad only w/o return filter           & 55.3 & \sgn{0.1} & 12.2 & 9.0  & 49.7 & 26.9 & 66.2 & 27.9 & 23.1 & 30.7 & \sdn{0.3} & 30.4--30.9 & \\
    \quad one level $\sigma=0.3$, w/o return filter & 56.9 & \sgn{1.7} & 12.8 & 10.6 & 49.3 & 26.2 & 65.2 & 32.4 & 24.8 & 31.6 & \sgn{0.6} & 31.2--32.0 & \\
    \bottomrule
  \end{tabular}}
\end{table*}

\paragraph{Baselines.} Our policy again outperforms every learned baseline, on
the training clips (55.2\% against 27.0\% for FDR and 23.9\% for SuperMimic) and zero-shot (31.0\% against 15.2 and
11.8\%), with 15.5 GPU-hours of training against 33.9 for FDR and 14.1 for SuperMimic. SPIDER, which optimizes every
zero-shot clip, converts more of them (42.4\%; \cref{sec:supp_spider}), at 110 GPU-hours for these clips (324 GPU-seconds per
clip), about seven times our training compute, and it searches anew for every new clip. Our policy's advantage is therefore the conversion of many clips at a
small, amortized cost, not the highest success on every clip under every physical setting. On a second hand with
stronger fingers, our policy keeps its success at the real torque limits and outperforms SPIDER and FDR there
(\cref{sec:supp_sharpa}).

\paragraph{Ablations.} Every whole group of components still matters: removing one lowers zero-shot success by 4.6 to 8.8
points, with all three seeds of every such row below those of Ours
(31.0--31.1\%), and success on the training clips by 4.0 to 10.8 points; the two noise variants stay within 0.6 points of Ours.
As at 4$\times$, beyond the contact geometry, object information matters for tracking success in the policy, not in the
world model: replacing only the policy's object
inputs lowers zero-shot success by 9.2 points, about as much as replacing those of both networks (8.8), whereas
replacing only the world model's leaves it unchanged (31.7\%, seeds 30.0--33.3\%). These inputs and the contact
geometry matter most. The gaps are smaller than at
4$\times$, since our policy itself reaches only 31.0\% at 1$\times$; the exploration design, for instance, contributes
4.6 points here, against 12.1 points at 4$\times$.

\paragraph{Training data.} Zero-shot success rises with every doubling of the data, from 23.8\% to
26.1, 29.1 and 31.0\% (\cref{tab:supp_ego}), and success on all 863 training clips of the full data from 48.3\% to
51.0, 53.1 and 55.2\%.
Adding EgoHOR raises the EgoHOR zero-shot sets about as much as at 4$\times$ ($+7.6$ points, from 21.0 to 28.6\%;
\cref{tab:supp_ego}); with all public data, it hardly changes the success on the public training clips (66.5 and
65.3\%). With a quarter
of the public data, however, it changes the public zero-shot sets only slightly (27.0 and 27.4\%); there, the weak
fingers rather than the coverage of the data presumably limit success.

\paragraph{Smoothness.} FDR's finger targets carry about 8 times as much content above \qty{5}{\hertz} as those of
our policy, and its wrist rotation targets and palm 2.3 to 2.5 times as much (\cref{tab:quality}), against about 5
and 2.5 times at 4$\times$ (\cref{sec:supp_ablation}).

\section{Transfer to a Second Robot Hand}
\label{sec:supp_sharpa}

This section gives the full setup (\cref{sec:supp_sharpa_setup}), per-set comparison with the XHAND
(\cref{sec:supp_sharpa_xhand}) and baselines (\cref{sec:supp_sharpa_base}), and ablations (\cref{sec:supp_sharpa_abl})
for the second robot hand summarized in \cref{sec:exp_sharpa}. With the same data and configuration, the method reaches
high success on the second hand also at its real torque limits, where it outperforms SPIDER and FDR, and every evaluated
group of its components still matters.

\subsection{Setup}
\label{sec:supp_sharpa_setup}

\begin{figure*}[!t]
  \centering
  \includegraphics[width=\textwidth]{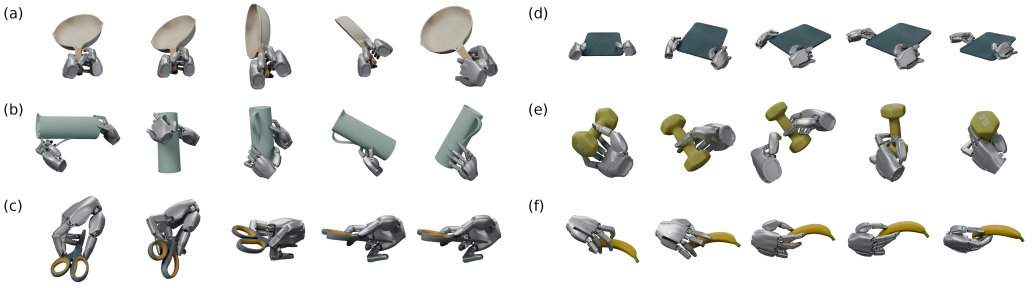}
  \caption{\textbf{Zero-shot tracking with the Sharpa at 1$\times$ torque} on the clips and moments of \cref{fig:gallery}
  (first seed; the banana is seen from the other side).}
  \label{fig:supp_sharpa_gallery}
\end{figure*}

The second hand is the Sharpa Wave, which has 22 finger joints instead of the XHAND's 12. Its references are
retargeted from the same human clips with the same kinematic retargeting, so the training data, the evaluation clips
and the configuration are those of the main experiments; only the robot hand differs, and with it the
hand-specific constants: each hand has 28 actuated joints instead of 18, which sets the size of the policy's output and of
$Q_h$ in \cref{eq:mirror}, and its own finger groups and torque limits; the left hand's inputs are mirrored as for the
XHAND. We train three seeds for 40k
iterations, with its finger torques limited to four times its real limits (4$\times$), as in our main experiments, and
to its real limits (1$\times$), and evaluate them as our main runs (\cref{tab:sharpa}). \Cref{fig:supp_sharpa_gallery}
shows the Sharpa at 1$\times$ on the clips of \cref{fig:gallery}.

\begin{table*}[!t]
  \centering
  \footnotesize
  \setlength{\tabcolsep}{4pt}
  \caption{\textbf{The Sharpa and the XHAND} with the same data, evaluation clips and configuration: SR$_{\text{MT}}$ (\%)
  on the training clips and per zero-shot set. XHAND rows: our main runs (\cref{tab:abl_sets,tab:sets_t1}). At
  1$\times$, the baselines and the ablation rows of \cref{tab:ablation} on the Sharpa (FDR for 600M steps); red: drop
  relative to Ours on the Sharpa at 1$\times$. Seeds: range of the zero-shot mean (--: one run); bold: best Train and
  Mean per torque. The row without the reward and training design is not run (exceeds the GPU memory).}
  \label{tab:sharpa}\label{tab:sharpa_abl}
  \fitcol{%
  \begin{tabular}{@{}lr@{\hspace{1pt}}lrrrrrrrr@{\hspace{1pt}}lr@{}}
    \toprule
    & & & \multicolumn{7}{c}{Zero-shot set (-ZS)} & & & \\
    \cmidrule(lr){4-10}
    & \multicolumn{2}{c}{Train} & HOT3D & DexYCB & OakInk2 & OakInk2-1H & TACO & EgoStd & GenHE & \multicolumn{2}{c}{Mean} & Seeds \\
    \midrule
    \multicolumn{13}{@{}l}{\textit{4$\times$ torque}} \\
    Sharpa, Ours                               & \textbf{80.7} & & 65.7 & 72.0 & 74.2 & 64.8 & 89.3 & 71.7 & 64.4 & \textbf{71.7} & & 71.1--72.8 \\
    XHAND, Ours                                & 69.4 & & 40.4 & 35.2 & 62.2 & 44.6 & 80.2 & 51.3 & 43.2 & 51.0 & & 49.6--53.5 \\
    \midrule
    \multicolumn{13}{@{}l}{\textit{1$\times$ torque}} \\
    Sharpa, Ours                               & \textbf{81.3} & & 63.8 & 67.8 & 73.0 & 63.2 & 89.4 & 69.9 & 64.8 & \textbf{70.3} & & 69.0--71.0 \\
    XHAND, Ours                                & 55.2 & & 13.5 & 10.1 & 48.4 & 25.3 & 62.9 & 31.2 & 26.0 & 31.0 & & 31.0--31.1 \\
    \addlinespace
    \multicolumn{13}{@{}l}{\textit{Sharpa, baselines}} \\
    \quad SPIDER~\citesupp{pan2025spider}       & --   & & 46.4 & 16.7 & 77.6 & 77.6 & 76.4 & 30.4 & 39.1 & 52.0 & & -- \\
    \quad SuperMimic~\citesupp{liao2026supermimic} & 23.8 & & 0.2  & 2.3  & 27.4 & 19.9 & 41.4 & 12.9 & 8.6  & 16.1 & & 14.9--17.1 \\
    \quad FDR~\citesupp{lee2026fdr}             & 26.9 & & 0.1  & 10.0 & 29.2 & 10.8 & 51.6 & 7.8  & 9.3  & 17.0 & & 15.6--17.7 \\
    \addlinespace
    \multicolumn{13}{@{}l}{\textit{Sharpa, Ours without}} \\
    \quad contact geometry $\phi$ (WM)          & 60.8 & \sdn{20.5} & 22.2 & 47.6 & 56.7 & 37.9 & 84.5 & 46.4 & 38.4 & 47.7 & \sdn{22.6} & 46.7--48.5 \\
    \quad object, scene inputs (WM, $\pi$)      & 72.5 & \sdn{8.8} & 23.9 & 45.4 & 57.3 & 35.5 & 85.6 & 54.9 & 39.2 & 48.8 & \sdn{21.5} & 48.6--49.0 \\
    \quad object, scene inputs (WM)             & 79.7 & \sdn{1.6} & 63.3 & 66.4 & 72.1 & 62.3 & 90.6 & 70.9 & 63.6 & 69.9 & \sdn{0.4} & 69.3--70.6 \\
    \quad object inputs ($\pi$)                 & 72.1 & \sdn{9.2} & 27.0 & 43.4 & 59.3 & 34.2 & 88.6 & 57.6 & 37.9 & 49.7 & \sdn{20.6} & 49.2--50.4 \\
    \quad two-hand structure                    & 78.0 & \sdn{3.3} & 47.6 & 56.1 & 69.8 & 66.4 & 91.4 & 64.9 & 55.1 & 64.5 & \sdn{5.8} & 63.6--65.2 \\
    \quad reward, training design               & --   & & --   & --   & --   & --   & --   & --   & --   & --   & & -- \\
    \quad exploration design                    & 56.3 & \sdn{25.0} & 13.5 & 18.4 & 46.3 & 45.2 & 70.3 & 40.4 & 28.2 & 37.5 & \sdn{32.8} & 34.4--41.4 \\
    \bottomrule
  \end{tabular}}
\end{table*}

\subsection{Comparison with the XHAND}
\label{sec:supp_sharpa_xhand}

With the same data, configuration and training, the two hands differ in two ways. At 4$\times$, where both have ample
finger torque, the Sharpa is already ahead: it tracks 71.7\% of the zero-shot clips and 80.7\% of the training clips,
against 51.0 and 69.4\% for the XHAND, presumably owing to its additional finger joints and its kinematics closer to
the human hand. From 4$\times$ to the real torque limits, the XHAND loses 20.0 points of zero-shot success (to 31.0\%),
whereas the Sharpa loses only 1.4 (to 70.3\%), presumably because its proximal finger joints are stronger (1.9 to
\qty{3.3}{N.m} against \qty{1.1}{N.m}). The method thus attains high success under real torque
limits on a hand with sufficient finger torque, whereas on the XHAND its success at 1$\times$ is limited by the finger
torque.

\subsection{Baselines at the real torque limits}
\label{sec:supp_sharpa_base}

At the real torque limits, the Sharpa also lets our policy outperform per-clip optimization. SPIDER, run on the Sharpa
with the same physics as our policy (\cref{sec:supp_spider}), converts 52.0\% of the zero-shot clips, against 70.3\%
for our policy, and needs about 150 GPU-hours for them, almost nine times our training (17.0). On the XHAND at
1$\times$, SPIDER is ahead (42.4 \vs 31.0\%; \cref{tab:sets_t1}); on the Sharpa, with the same hand for both methods,
our policy leads by 18.3 points (95\% interval $[13.8, 22.8]$ with objects resampled, as in \cref{tab:supp_ci}), and
SPIDER is ahead only on OakInk2-ZS and OakInk2-ZS-1H. FDR, trained on the Sharpa for 600M steps (about 40 GPU-hours), converts 17.0\% of
the zero-shot clips and 26.9\% of the training clips, about as many as on the XHAND (15.2 and 27.0\%): the stronger
fingers, which raise our policy from 31.0 to 70.3\% zero-shot, do not help it, and our policy leads by 53.3 points
($[48.6, 57.5]$).

\subsection{Ablations}
\label{sec:supp_sharpa_abl}

We repeat the ablation rows of \cref{tab:ablation} on the Sharpa at 1$\times$
(\cref{tab:sharpa_abl}). Every evaluated group of components still matters: removing one lowers zero-shot success by 5.8 to 32.8
points, with all three seeds of every such row below those of Ours, except replacing the world model's object and scene
inputs alone ($-0.4$ points, within the seed range of Ours), as on the XHAND ($+0.9$ points at 4$\times$ and $+0.7$ at 1$\times$). The ranking agrees
with that on the XHAND: beyond the contact geometry, object information matters for tracking success in the policy, not
in the world model, and the contact geometry and
the exploration design matter most. The row without the reward and training design is not run, since with the Sharpa
it does not fit in \qty{48}{GB} of GPU memory: it imagines 32 steps instead of 24 with the full contact geometry and
point-cloud encoding.

\ifdefined\nolinenumbers\nolinenumbers\fi
\section*{References}
{\small
\begin{list}{}{\setlength{\labelwidth}{2.1em}\setlength{\labelsep}{0.4em}\setlength{\leftmargin}{2.5em}\setlength{\itemsep}{1pt}\setlength{\parsep}{0pt}\setlength{\topsep}{0pt}\setlength{\partopsep}{0pt}}
\item[{[S1]}] \hypertarget{sref:genrobot2026genhumanego}{}GenRobot. Gen-HumanEgo: Egocentric human demonstration dataset. Hugging Face Dataset, 2026.
\item[{[S2]}] \hypertarget{sref:jiang2025dexmimicgen}{}Zhenyu Jiang, Yuqi Xie, Kevin Lin, Zhenjia Xu, Weikang Wan, Ajay Mandlekar, Linxi Fan, and Yuke Zhu. DexMimicGen: Automated data generation for bimanual dexterous manipulation via imitation learning. In {\em IEEE Int. Conf. Robot. Autom.}, 2025.
\item[{[S3]}] \hypertarget{sref:lee2026fdr}{}Kyungmin Lee, Sibeen Kim, Dongyoon Hwang, Yoonsang Oh, Donghu Kim, Youngdo Lee, I~Made~Aswin Nahrendra, Jaegul Choo, and Hojoon Lee. FlashDexRetarget: Accelerating dexterous manipulation data generation through multi-motion retargeting. {\em arXiv preprint arXiv:2610.01849}, 2026.
\item[{[S4]}] \hypertarget{sref:liao2026supermimic}{}Qiayuan Liao, Yiyang Shao, Junfeng Long, Kevin Zakka, Roberto Horowitz, Pieter Abbeel, and Koushil Sreenath. SuperMimic: Scaling humanoid whole-body control with one differentiable world model. {\em arXiv preprint arXiv:2610.28480}, 2026.
\item[{[S5]}] \hypertarget{sref:lightwheel2026egosuite}{}Lightwheel. EgoSuite-Open100K: A fully annotated open egocentric human dataset. \url{https://huggingface.co/collections/LightwheelAI/egosuite-open100k}, 2026. Dataset.
\item[{[S6]}] \hypertarget{sref:mandi2025dexmachina}{}Zhao Mandi, Yifan Hou, Dieter Fox, Yashraj Narang, Ajay Mandlekar, and Shuran Song. DexMachina: Functional retargeting for bimanual dexterous manipulation. In {\em Int. Conf. Mach. Learn.}, 2026.
\item[{[S7]}] \hypertarget{sref:paliwal2026doasido}{}Bhawna Paliwal, Haritheja Etukuru, William Liang, Pieter Abbeel, Nur Muhammad~Mahi Shafiullah, and Jitendra Malik. Do as I do: Dexterous manipulation data from everyday human videos. {\em arXiv preprint arXiv:2606.19333}, 2026.
\item[{[S8]}] \hypertarget{sref:pan2025spider}{}Chaoyi Pan, Changhao Wang, Haozhi Qi, Zixi Liu, Homanga Bharadhwaj, Akash Sharma, Tingfan Wu, Guanya Shi, Jitendra Malik, and Francois Hogan. SPIDER: Scalable physics-informed dexterous retargeting. {\em arXiv preprint arXiv:2511.09484}, 2025.
\item[{[S9]}] \hypertarget{sref:wei2022coacd}{}Xinyue Wei, Minghua Liu, Zhan Ling, and Hao Su. Approximate convex decomposition for 3D meshes with collision-aware concavity and tree search. {\em ACM Trans. Graph.}, 2022.
\item[{[S10]}] \hypertarget{sref:zhu2026chord}{}Xinghao Zhu, Zixi Liu, Shalin Jain, Chenran Li, Milad Noori, Michael~Andres Lin, Huihua Zhao, John Welsh, Mrinal Verghese, Wei Liu, Tingwu Wang, Xingye Da, Zhengyi Luo, Vishal Kulkarni, Naema Bhatti, Yuke Zhu, Linxi Fan, Bowen Wen, Danfei Xu, Soha Pouya, and Yan Chang. Learning dexterous manipulation using contact wrench guidance from human demonstration. {\em arXiv preprint arXiv:2607.00033}, 2026.
\end{list}}

\end{document}